%% file: paper_template.tex
\documentclass[conference]{IEEEtran}

\let\labelindent\relax

\usepackage{times}
\usepackage{multicol}
\usepackage[bookmarks=true]{hyperref}
\usepackage[numbers, sort&compress]{natbib}
\usepackage{siunitx}
\usepackage{enumitem}
\usepackage{booktabs}
\usepackage{graphicx}
\usepackage{subcaption}
\usepackage{titletoc}
\usepackage{amssymb}
\usepackage{amsmath}
\usepackage{cleveref}

\newcommand{\weburl}{\href{https://mujocoplayground.github.io/}{ \texttt{mujocoplayground.github.io}}}

\begin{document}

\title{MuJoCo Playground}

\author{
Kevin Zakka$^{*,1}$, Baruch Tabanpour$^{*,2}$, Qiayuan Liao$^{*,1}$, Mustafa Haiderbhai$^{*,3}$, Samuel Holt$^{*,4}$, \\Jing Yuan Luo$^{2}$, Arthur Allshire$^{1}$, Erik Frey$^{2}$, Koushil Sreenath$^{1}$, Lueder A. Kahrs$^{3}$, \\Carmelo Sferrazza$^{{\dagger},1}$, Yuval Tassa$^{{\dagger},2}$, and Pieter Abbeel$^{{\dagger},1}$ \\ \\ 
$^1$UC Berkeley \qquad $^2$Google DeepMind \qquad $^3$University of Toronto \qquad $^4$University of Cambridge \\
$^*$Equal contributions \qquad $^\dagger$Equal advising
}

\makeatletter
\let\@oldmaketitle\@maketitle%
\renewcommand{\@maketitle}{\@oldmaketitle%
    \centering
    \vspace{1em}
    \includegraphics[width=0.8\textwidth]{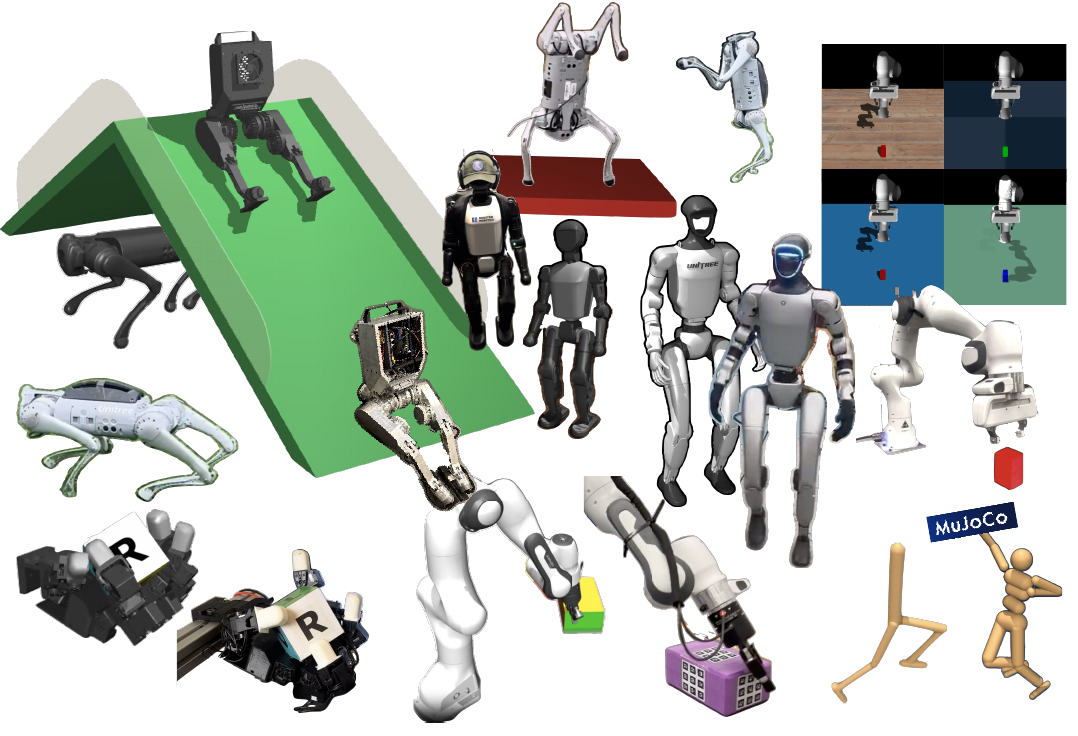}
    \captionof{figure}{
        \small{A cartoon of MuJoCo Playground’s diverse environments that were successfully transferred to real hardware, including Berkeley Humanoid, Unitree Go1 and G1, LEAP hand and Franka Arm.}
    }
    \label{fig:main_figure}
    }
\makeatother

\maketitle
\addtocounter{figure}{-1}

\begin{abstract}
We introduce \href{https://playground.mujoco.org/}{MuJoCo Playground}, a fully open-source framework for robot learning built with MJX, with the express goal of streamlining simulation, training, and sim-to-real transfer onto robots. With a simple \texttt{pip install playground}, researchers can train policies in minutes on a single GPU. \mbox{Playground} supports diverse robotic platforms, including quadrupeds, humanoids, dexterous hands, and robotic arms, enabling zero-shot sim-to-real transfer from both state and pixel inputs. This is achieved through an integrated stack comprising a physics engine, batch renderer, and training environments. Along with video results, the entire framework is freely available at \weburl.
\end{abstract}

\IEEEpeerreviewmaketitle

\input{sections/1_intro}
\input{sections/2_environments}
\input{sections/3_madrona}

\input{sections/4_results}
\input{sections/5_relatedwork}
\input{sections/8_limitations}
\input{sections/6_conclusion}
\input{sections/7_ack}

\bibliographystyle{plainnat}
\bibliography{references}

\clearpage
\onecolumn
\appendices

\makeatletter
\renewcommand{\thesection}{\Alph{section}} %
\renewcommand{\thesubsection}{\thesection.\arabic{subsection}} %
\renewcommand{\@seccntformat}[1]{%
  \csname the#1\endcsname.\hspace{1em}} %
\makeatother

\begin{center}
{\huge \textbf{Appendix}}    
\end{center}

\addtocontents{toc}{\protect\setcounter{tocdepth}{2}} %
\startcontents
\printcontents{}{1}{\textbf{Table of Contents}\vskip3pt\hrule\vskip5pt} \vskip3pt\hrule\vskip5pt

\input{appendix/dm_control.tex}
\input{appendix/locomotion.tex}
\input{appendix/manipulation.tex}
\input{appendix/madrona.tex}
\input{appendix/rl_details.tex}

\end{document}

%% file: sections/1_intro.tex
\section{Introduction}

Reinforcement learning (RL)~\cite{kober2013reinforcement} with subsequent transfer to hardware (sim-to-real)~\cite{zhao2020sim}, is emerging as a leading paradigm in modern robotics~\cite{kaufmann2023champion, lee2019robust, miki2022learning}. The benefits of simulation are obvious -- safety and cheap data. The recipe involves four steps:
\begin{enumerate}[leftmargin=0.8cm, rightmargin=0.4cm, topsep=0em, itemsep=2pt] 
    \item Create a simulated environment that matches the~real~world.
    \item Encode desired robot behavior with a reward function.
    \item Train a policy in simulation.
    \item Deploy to the robot. 
\end{enumerate}

The key enabler of this approach is a simulator that is realistic, convenient, and fast. 

The realism requirement is self-evident, the ``digital twin'' of step 1 demands a minimal level of fidelity~\cite{zhao2020sim}. Convenience and usability are equally critical, streamlining the creation, modification, composition, and characterization (system identification) of simulated robots.

The importance of speed is less obvious -- why does it matter if training takes ten minutes or ten hours? The answer lies in reward design (step 2), which cannot be easily automated: what the robot \emph{ought} to do is an expression of human preference. Even if reward design is semi-automated~\cite{ma2023eureka}, the process remains iterative: RL excels at finding policies that obtain reward, but the resulting behavior is often irregular \href{https://www.youtube.com/watch?v=EI3gcbDUNiM&t=257s}{in unexpected ways}.
Since steps 2 and 3 (and occasionally step 4) must be repeated~\cite{chebotar2019closing}, \mbox{\emph{time-to-robot}} becomes critical: the time from when you ask the robot to do something until you see what it thinks you meant.

RL is computationally intensive, requiring an enormous number of agent-environment interactions to train effective policies~\cite{ibarz2021train}. GPU-based simulation can significantly accelerate this process for two key reasons. First, the median GPU is far more powerful than the median CPU~\cite{wang2020benchmarking}, and while high core-count CPUs exist, they are uncommon. Second, by keeping the entire agent-environment loop on device, we can harness the high-throughput, highly parallel architecture~\cite{makoviychuk2021isaac, freeman2021brax}. This is especially true for \emph{on-policy} RL~\cite{schulman2017proximal, andrychowicz2020matters}, which employs GPU-friendly, wide-batch operations. Locomotion and manipulation tasks which previously required days of training on multi-host setups~\cite{andrychowicz2020learning,tan2018sim}, can now be solved within minutes or hours on a single GPU~\cite{rudin2022learning, handa2023dextreme}.

With this work, we aim to further advance and make sim-to-real robot learning even more accessible. We introduce MuJoCo Playground, a fully open-source framework for robot learning designed for rapid iteration and deployment of sim-to-real reinforcement learning policies. We build upon MuJoCo XLA~\cite{mujocoxla} (MJX), a JAX-based branch of the MuJoCo physics engine that runs on GPU, enabling training directly on device. Besides physics and learning, we leverage the open-source nature of our ecosystem to incorporate on-device rendering through the Madrona batch renderer~\cite{shacklett2023extensible}, facilitating training of vision-based policies end-to-end, without teacher-student distillation~\cite{agarwal2023legged}. With a straightforward installation process (\texttt{pip install playground}) and cross-platform support, users can quickly train policies on a single GPU. The entire pipeline---from environment setup to policy optimization---can be executed in a single Colab notebook, with most tasks requiring only minutes of training~time.

MuJoCo Playground’s lightweight implementation greatly simplifies sim-to-real deployment, transforming it into an interactive process where users can quickly tweak parameters to refine robot behavior. In our experiments, we deployed both state- and vision-based policies across six robotic platforms in less than eight weeks. We hope that MuJoCo Playground becomes a valuable resource for the robotics community and  expect it to continue building on MuJoCo's thriving open-source ecosystem.

\noindent Our work makes three main contributions:
\begin{enumerate}
    \item We develop a comprehensive suite of robotic environments using MJX \cite{mujocoxla}, demonstrating sim-to-real transfer across diverse platforms including quadrupeds, humanoids, dexterous hands, and robot arms.
    \item We integrate the open-source Madrona batch GPU renderer \cite{shacklett2023extensible} to enable end-to-end vision-based policy training on a single GPU device, achieving zero-shot transfer on manipulation tasks.
    \item We provide a complete, reproducible training pipeline with notebooks, hyperparameters, and training curves, enabling rapid iteration between simulation and real-world deployment.
\end{enumerate}

\begin{figure*}[t!]
    \centering
    \includegraphics[width=1\linewidth]{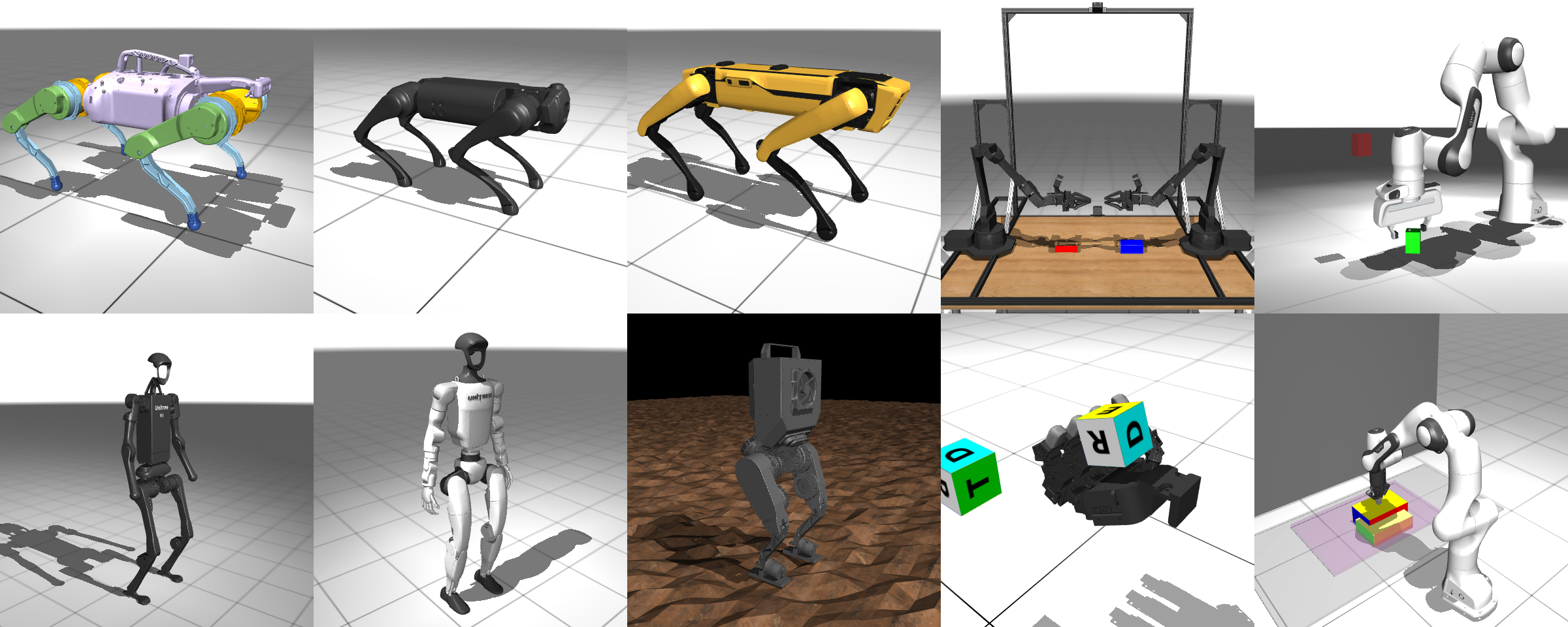}
    \caption{\small A preview of locomotion and manipulation environments available in MuJoCo Playground.}
    \label{fig:env_grid}
\end{figure*}

%% file: sections/2_environments.tex
\section{Environments}

MuJoCo Playground contains environments in 3 main categories: DeepMind (DM) Control Suite,  Locomotion, and Manipulation, which we briefly describe in this section. Locomotion and manipulation environments are tailored to robotic use-cases and we show zero-shot sim-to-real transfer in many of the available environments. Playground directly utilizes MuJoCo Menagerie \cite{menagerie2022github} which offers a suite of robot assets and configurations tailored to run in MuJoCo.

\subsection{DM Control Suite}

The majority of RL environments from \cite{tassa2018deepmind} are re-implemented in MJX, and serve as entry-level tasks to familiarize users with MuJoCo Playground (\Cref{fig:dmc_grid}).

\subsection{Locomotion}

Locomotion environments in MuJoCo Playground are implemented for multiple quadrupeds and bipeds (\Cref{fig:env_grid} left). The quadrupeds include the Unitree Go1, Boston Dynamics Spot, and Google Barkour \cite{caluwaerts2023barkour}, while the humanoids include the Berkeley Humanoid \cite{liao2024berkeley}, Unitree H1 and G1, Booster T1, and the Robotis OP3. For each robot embodiment, we implement a joystick environment that learns to track a velocity command consisting of base linear velocities in both the forward and lateral directions, as well as a desired yaw rate. On the Unitree Go1, we additionally implement fall recovery and handstand environments. A complete list of locomotion environments is provided in \Cref{tab:locomotion_envs} in the appendix.

We demonstrate sim-to-real transfer in two main sets of experiments. First, on the Unitree Go1, we deploy joystick, fall recovery, and handstand policies. Second, we demonstrate joystick-based locomotion on the Berkeley Humanoid, the Unitree G1, and the Booster T1. More details on these sim-to-real experiments can be found in \Cref{sec:results_sim2real_locomotion}.

\subsection{Manipulation}

Manipulation environments in MuJoCo Playground are implemented for both prehensile and non-prehensile tasks (\Cref{fig:env_grid} right). With the Leap Hand \cite{shaw2023leaphand} robot, we demonstrate contact-rich dexterous re-orientation of a block. Using the Franka Emika Panda and Robotiq gripper, we show re-orientation of a yoga block using high frequency torque control. We implement a simple vision-based pick-cube environment on a Franka arm using the Madrona batch renderer. A few additional environments, such as bi-arm peg-insertion with the Aloha robot \cite{aldaco2024aloha}, are also available. We refer to \Cref{tab:manipulation_envs} in the appendix for a full set of environments.

We demonstrate sim-to-real transfer on the Leap Hand and Franka arm robots, including an environment trained from vision for the pick-cube task. More details on the sim-to-real experiments are available in \Cref{sec:results_sim2real_manipulation}.

%% file: sections/3_madrona.tex
\section{Batch Rendering with Madrona}
\label{sec:madrona}

MuJoCo Playground enables vision-based environments through an integration of MJX with Madrona \cite{shacklett23madrona}. Madrona is a GPU-based entity-component-system (ECS), which contains GPU implementations of high throughput rendering \cite{rosenzweig24madronarenderer}. Madrona provides two rendering backends: a software-based batch ray tracer written in CUDA (used for the experiments in this work) and a Vulkan-based rasterizer. The raytracing backend supports features including complex lighting scenarios, shadows, textures, and geometry materials. See \Cref{fig:madrona_highres} for examples of rendered images using the batch ray tracer. Some features such as deformable materials, moving lights, and terrain height fields will be added in the future.

The Madrona Batch Renderer is integrated with MJX through low-level JAX \cite{jax2018github} primitives that connect to the initialization and render functions exposed by Madrona. These JAX primitives allow for Madrona to interact seamlessly with JAX transformations such as \textit{jit} and \textit{vmap}. Mujoco Playground provides two examples: (\texttt{cartpole-balance} and \texttt{PandaPickCubeCartesian}) to showcase the implementation of vision-based environments and training of vision-based policies.

The Madrona MJX integration also supports customization of each environment instance, allowing for domain randomization~\cite{domainrand2017} of visual properties such as geometry size, color, lighting conditions, and camera pose. These randomizations play a crucial role in the sim-to-real transfer of vision-based policies, which we discuss more in \Cref{sec:results_pickcube_pixels}.

%% file: sections/4_results.tex
\section{Results}
\label{sec:results}

\begin{figure*}[t]
    \centering
    \includegraphics[width=1\linewidth]{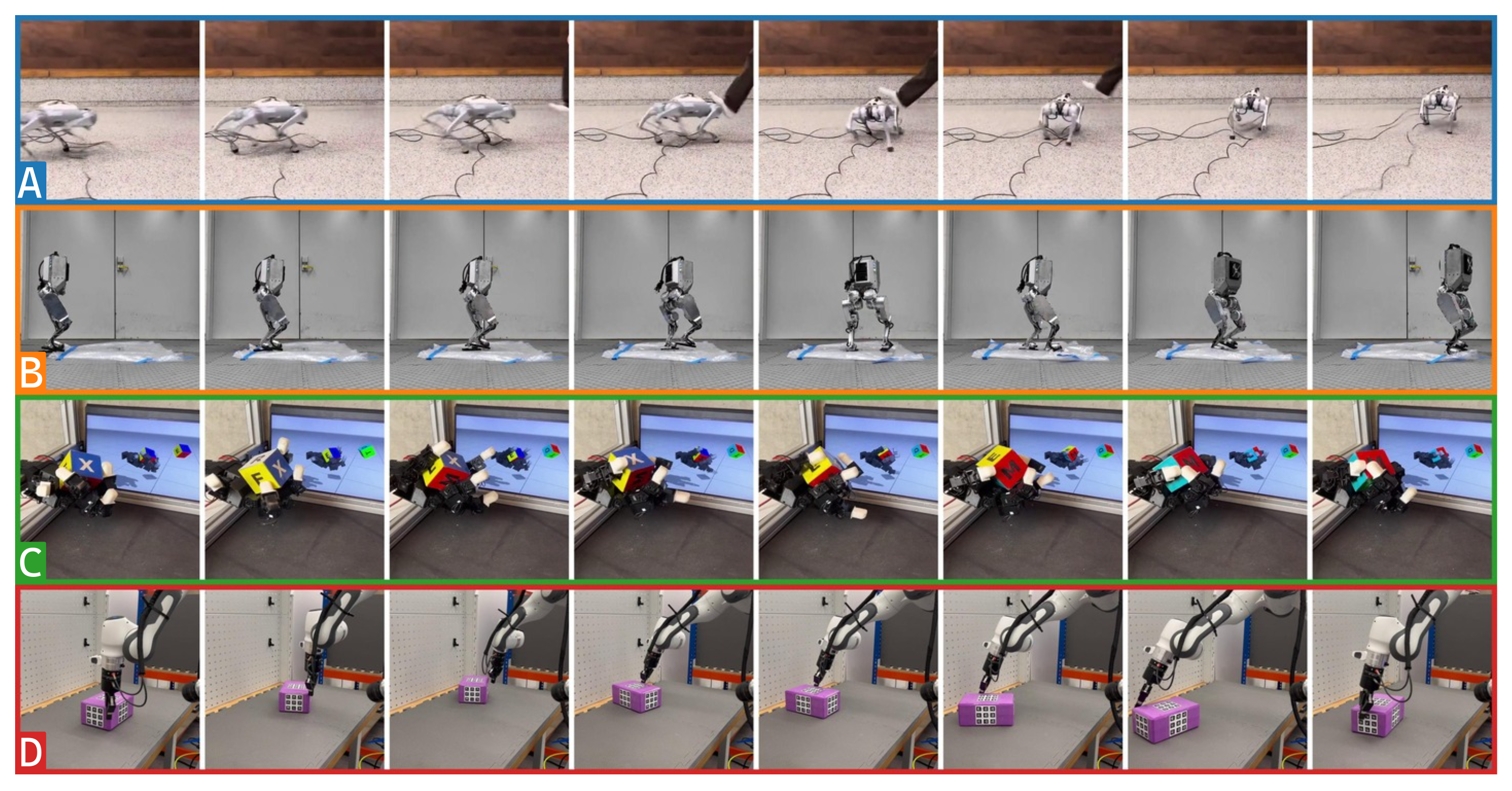}
    \caption{\small Footage from four of our deployed policies. a) Go1 joystick policy recovering from a kick while travelling at $\sim$ 2m/s, b) Berkeley humanoid joystick policy tracking an angular velocity command on a slippery surface. c) In-Hand Cube Reorientation transitioning between two target poses. d) Non-prehensile policy issuing torque commands to rotate a block by 180 degrees.}
    \label{fig:action_reel}
\end{figure*}

In this section, we report RL and sim-to-real results for environments in MuJoCo Playground. Sim-to-real experiments (see some examples in \Cref{fig:action_reel}) are performed for locomotion and manipulation environments from both proprioceptive state and from vision.
We briefly discuss RL training on different hardware devices and RL libraries.

\subsection{DM Control Suite}

We train state-based policies for all available tasks, with most environments training in under 10 minutes on a single GPU device. More details on the training process can be found in \Cref{sec:appendix_dm_control_curves}. All available environments in the MJX port of the DM Control Suite, including any modifications, are detailed in \Cref{sec:appendix_dm_control_envs}. 

Using the batch renderer, we also implement pixel-based observations for the CartpoleBalance environment. These observations are generated on the GPU, allowing us to keep physics, rendering, and training entirely on-device. Although other DM Control Suite environments can also be rendered with Madrona, we demonstrate end-to-end RL training on only one task, leaving a more comprehensive exploration for future work. \Cref{appendix:MadronaMJXBenchmark} provides more information on how CartpoleBalance was modified and trained for pixel observations.

\begin{figure}[h!]
    \centering
    \includegraphics[width=1\linewidth]{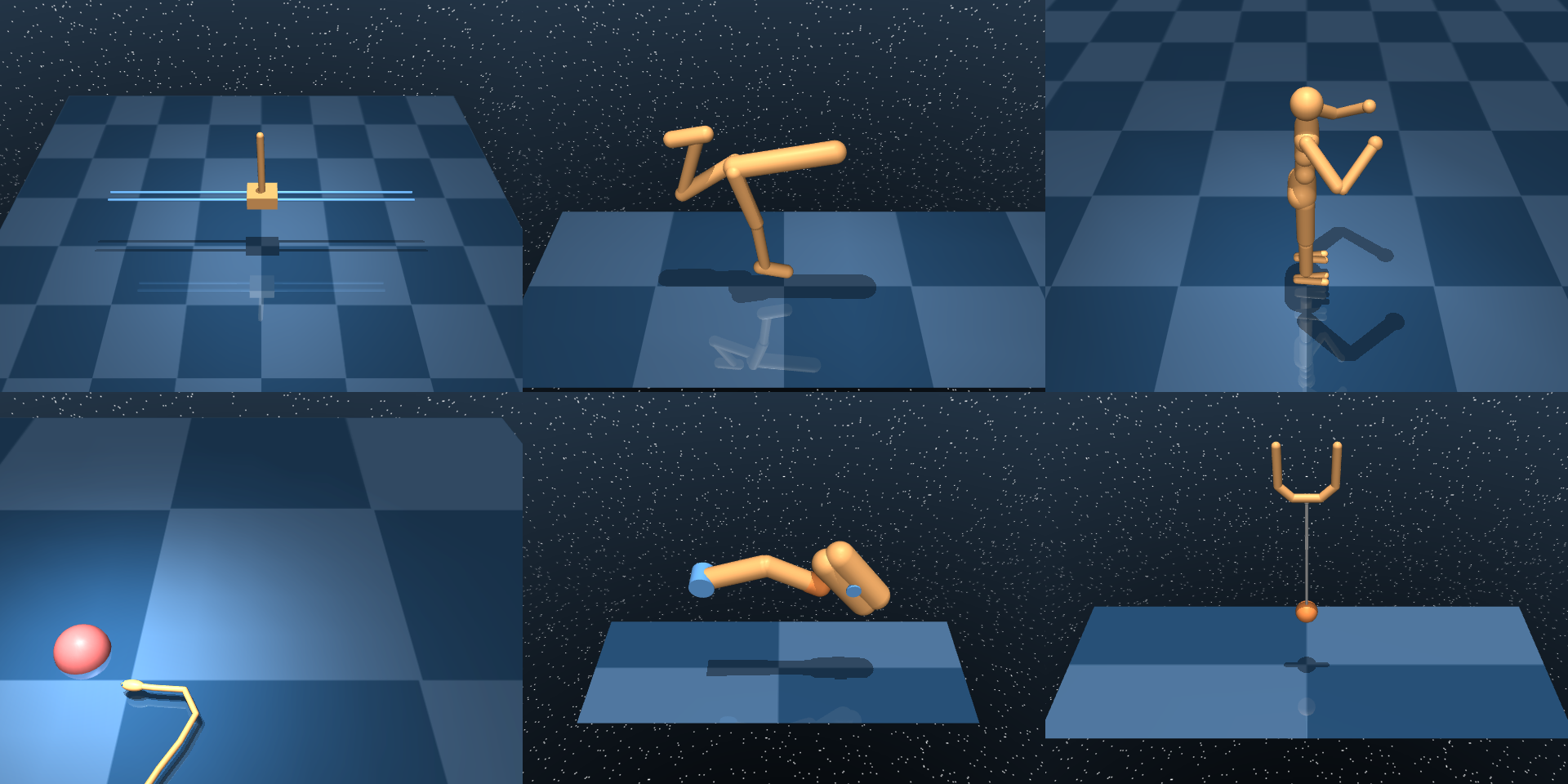}
    \caption{\small Several DM Control Suite environments.}
    \label{fig:dmc_grid}
\end{figure}

\subsection{Locomotion}
\label{sec:results_sim2real_locomotion}

We present sim-to-real locomotion results on both a quadruped (Unitree Go1) and three humanoid platforms (Berkeley Humanoid,Unitree G1, and Booster T1). Further details on the MDP formulation, including rewards, observation spaces, and action spaces, are provided in \Cref{sec:appendix_locomotion}.

\subsubsection{Quadruped Locomotion}

\begin{figure}[h]
    \centering
    \includegraphics[width=1\linewidth]{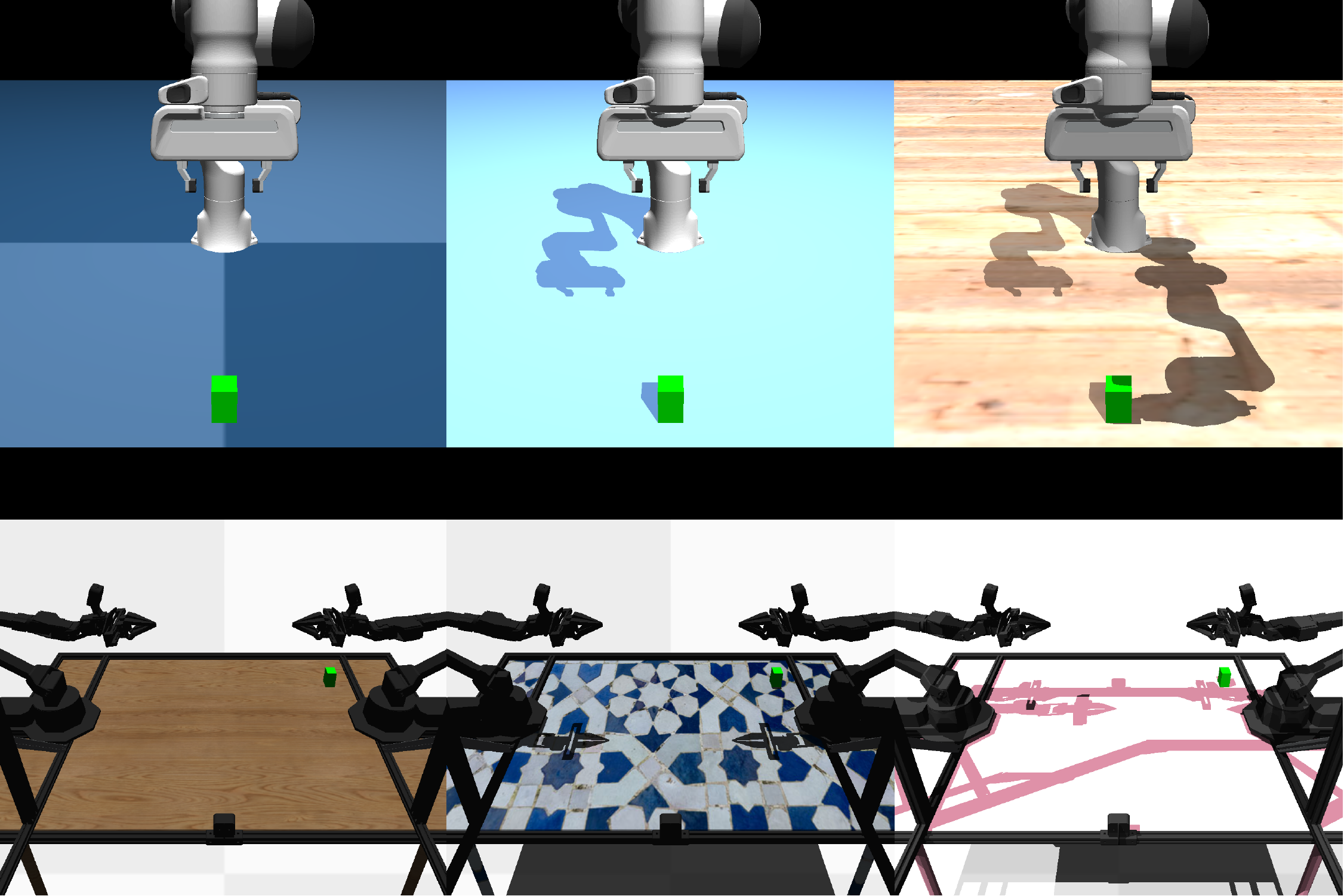}
    \caption{\small Sample renders from the Madrona batch renderer for the Panda and Aloha environments. Left-most images are the original environments. The remaining images highlight the the support for lighting, shadows, textures, and colors, including the ability to domain randomize these parameters during training.}
    \label{fig:madrona_highres}
\end{figure}

\paragraph{Task definition} We implement a joystick locomotion task as in \cite{rudin2022learning, Ji_2022}, where the command is specified by three values indicating the desired forward velocity, lateral velocity, and turning rate of the robot’s root body. Additionally, we design policies for handstand and footstand tasks, in which the robot balances on the front or hind legs, respectively, while minimizing actuator torque. For fall recovery, we follow \cite{lee2019robust, smith2022legged}, enabling the robot to return to a stable ``home'' posture from arbitrary fallen configurations.

\paragraph{Hardware} We deploy on the \textit{Unitree Go1}, which is a quadruped robot with four legs, each possessing three degrees of freedom. Trained policies run on real-world outdoor terrain (grass and concrete) and indoor surfaces with different friction properties.

\paragraph{Training} We domain randomize for sensor noise, dynamics properties and task uncertainties. We firstly train the policy in flat ground with restricted command ranges within 5 minutes (2x RTX 4090). and finetune it in rough terrain with wider ranges. See \Cref{sec:appendix_locomotion} for more detail.

\paragraph{Results} All four policies (joystick, handstand, footstand, and fall recovery) transfer robustly from simulation to reality, coping with uneven terrain and moderate external perturbations without additional fine-tuning. Videos of these deployments are provided on our project website.

\subsubsection{Humanoid Locomotion}

\paragraph{Task definition} We implement the same joystick locomotion task as shown for the quadruped environment.

\paragraph{Hardware} We perform sim-to-real experiments on three different humanoid platforms: a) \textit{Berkeley Humanoid} \cite{liao2024berkeley}, a low-cost, lightweight bipedal robot with 6\,DoF per leg, b) \textit{Unitree G1}, a humanoid robot featuring 29\,DoF in total, and c) \textit{Booster T1}, a small-scale humanoid robot with 23\,Dof. All systems are evaluated in indoor environments, with slight variations in surface friction and ground compliance.

\paragraph{Training} We follow the domain randomization and finetuning strategies of the quadruped robot. Training on flat ground lasts under 15 minutes for the Berkeley Humanoid, and under 30 minutes for the Unitree G1 and the Booster T1 on two RTX 4090.

\paragraph{Results} We successfully deploy joystick-based locomotion on the Berkeley Humanoid, demonstrating robust tracking of velocity commands on surfaces ranging from rigid floors to soft and slippery terrains. On the Unitree G1 and Booster T1, our zero-shot policy similarly achieves stable walking and turning on standard indoor floors. Although minor tuning for each platform’s unique dynamics may further enhance performance, these results confirm that our approach generalizes across a range of legged robot morphologies.

\subsection{Manipulation}
\label{sec:results_sim2real_manipulation}

In this section, we present sim-to-real results for a broad range of manipulation tasks, including dexterous in-hand manipulation, non-prehensile manipulation, and vision-based grasping. These tasks illustrate Playground's ability to address a diverse segment of the manipulation spectrum and highlight its robust deployment in real-world settings.

\subsubsection{In-Hand Cube Reorientation}
\label{sec:in_hand_cube_reorientation}

\paragraph{Task definition} We implement an in-hand cube reorientation task using the low-cost, dexterous LEAP hand platform \cite{shaw2023leaphand}, closely following previous works on in-hand manipulation \cite{handa2023dextreme, andrychowicz2020learning}. The task involves reorienting a 7\,cm cube repeatedly from random initial poses to new target orientations in SE(3) without dropping it. Further task details are provided in \Cref{sec:appendix_leaphand_real}.

\paragraph{Hardware} We employ the same hardware configuration as in \cite{li2024_drop}, mounting the LEAP hand on an 80/20 frame with a 3D-printed bracket that tilts the palm downward by 20°. A single Intel RealSense D415 camera, positioned above the workspace, provides pose estimates of the cube via a pretrained detector \cite{handa2023dextreme}. Although occlusions can introduce observation noise, we leave multi-camera extensions to future work. The policy operates at 20\,Hz, which remains comfortably below the USB-Dynamixel control bandwidth.

\paragraph{Training} To promote sim-to-real transfer, we apply domain randomization on the robot parameters as well as cube mass and friction. We also include sensor noise, and we finetune with a progressive curriculum to increase both noisy pose estimates and action regularization. The policy trains within 30\,min on two RTX 4090 GPUs. Further training details are provided in \Cref{sec:appendix_leaphand_real}.

\paragraph{Results} As summarized in \Cref{tab:leaphandcubereorientationdetails}, our learned policy demonstrates early signs of robust in-hand reorientation with MuJoCo Playground. The most frequent failure occurs when the cube becomes wedged in the space present between the fingers and the palm of the LEAP hand, causing the policy to stall. Although less common, we also observe accidental interlocking of the index and thumb, attributed to physical flex in the low-cost hardware. Videos of real-world deployments can be found on our project page. We note that improved camera coverage and more accurate collision geometries could mitigate these edge-case failures, which we leave for future work.

\begin{table}[t]
  \centering
  \caption{\small{In-hand reorientation results on the LEAP hand over 10 trials, reporting the number of consecutive successful rotations before failure. The final two columns show the median and mean of the \#Rotations metric.}}
  \label{tab:leaphandcubereorientationdetails}
  \vspace{-5pt}
  \scriptsize
  \renewcommand{\arraystretch}{1.1}
  \setlength{\tabcolsep}{5pt}
  \resizebox{\columnwidth}{!}{%
    \begin{tabular}{l|cccccccccc|cc}
    \toprule
    & \multicolumn{10}{c|}{\textbf{Trial}} & \multicolumn{2}{c}{\textbf{Summary}} \\
    \cmidrule{2-13}
    & 1 & 2 & 3 & 4 & 5 & 6 & 7 & 8 & 9 & 10 & Median & Mean \\
    \midrule
    \# Rotations 
      & 3  & 27 & 8  & 2  & 15 & 3  & 4  & 1  & 3  & 5  
      & \textbf{3.5} 
      & \textbf{7.1} \\
    \bottomrule
    \end{tabular}
  }
  \vspace{-10pt}
\end{table}

\subsubsection{Non-Prehensile Block Reorientation}

\paragraph{Task definition}
We present a sim-to-real setup for non-prehensile reorientation of a yoga block on a commonly available Franka Emika Panda robot arm with a Robotiq gripper, achieving high zero-shot success. The task involves moving a yoga block from a random initial pose in the robot’s workspace to a fixed goal pose. A trial is deemed successful if the agent reorients the block within 3\,cm of the goal position and within 10° of the desired orientation.

\paragraph{Hardware}
The policy receives estimates of the block’s position and orientation from an open-source camera tracker \citep{artrackalvar}. We use direct high-frequency torque control at 200\,Hz, where the RL policy outputs motor torques for the arm’s seven joints (with the gripper closed). By learning to control torques rather than joint positions, the agent develops smooth, compliant behavior that transfers effectively to hardware, delivering superior performance even when direct torque control at high frequencies poses learning challenges~\citep{holt2024evolving}. This recipe, therefore, holds broad value for practitioners.

\paragraph{Training}
Robust zero-shot transfer is enabled by stochastic delays and progressive curriculum learning. Each training episode injects randomization into initial poses, joint positions, and velocities, while also imposing action and observation stochastic delays to mirror practical hardware latency. A simple curriculum gradually increases the block’s displacement and orientation range upon each success, preventing overfitting to easier conditions. Training takes 10 minutes on 16x A100 devices.

\paragraph{Results}
These techniques, combined with 200\,Hz direct torque control, produce a policy resilient to real-world perturbations. The agent reliably reorients the block on hardware with no additional fine-tuning as shown in \Cref{table:blockreorientfrankarobotiqrealresults}. Videos of real-world deployments are provided on our project website. Additional implementation details are given in \Cref{appendix:RealWorldNonprehensileReorientationPolicyPerformance}.

\begin{table}[!tb]
  \centering
  \caption{\small{Sim-to-real reorientation performance on the Franka Emika Panda robot, evaluated across 35 hardware trials. Each metric is reported as the median and mean (with a 95\% confidence interval). The success rate is bolded to highlight final task performance. The training was done on 16x A100 GPUs.}
  }
  \vspace{-10pt}
  \smallskip
  \resizebox{\columnwidth}{!}{
    \begin{tabular}{@{}l|c|c}
    \toprule
    Metric & Median & Mean ± 95\% Confidence Interval \\
    \midrule
    \textbf{Real Success (\%)} $\uparrow$ & \textbf{100} & \textbf{85.7 ± 12.2} \\
    Position Error (cm) $\downarrow$ & 1.95 & 5.28 ± 3.26 \\
    Rotation Error (°) $\downarrow$ & 1.72 & 3.32 ± 1.59 \\
    \bottomrule
    \end{tabular}
  }
  \vspace{-20pt}
  \label{table:blockreorientfrankarobotiqrealresults}
\end{table}

\subsubsection{Pick-Cube from Pixels}
\label{sec:results_pickcube_pixels}

\paragraph{Task definition}
We demonstrate sim-to-real transfer with pixel-based policies on a Franka Emika Panda robot. The robot must reliably grasp and lift a small 2\,\(\times\)\,2\,\(\times\)\,3\,cm block from a random location on the table and move it 10\,cm above the surface. The policy receives a \(64\times64\) RGB image as input and outputs a Cartesian command, which is processed by a closed-form inverse kinematics solution to yield joint commands. To simplify the task, we restrict the end-effector to a 2D Y-Z plane (while always pointing downward) and provide a binary jaw open/close action.

\paragraph{Hardware}
We use a Franka Emika Panda robot with a single Intel RealSense D435 camera mounted to capture top-down RGB images. The policy operates at 15\,Hz, and we run inference on an RTX 3090 GPU. Our setup ensures that the block starts within the field of view over a 20\,cm range along the y-axis.

\paragraph{Training}
To bridge the sim-to-real gap, we apply domain randomization across visual properties such as lighting, shadows, camera pose, and object colors. We also add random brightness post-processing, and introduce a stochastic gripping delay of up to 250\,ms.
We choose a reduced action dimension of three (Y-movement, Z-movement, and discrete jaw control) for training sample efficiency, but we have found that the task can also be solved in full Cartesian or joint space given additional camera perspectives and more training samples. Training in simulation takes ten minutes on a single RTX 4090.

\paragraph{Results}
Our policy achieves a 100\% success rate in 12 real-world trials, robustly grasping the block and lifting it clear of the table. It demonstrates resilience to moderate variations in lighting and minor camera shaking, as shown in the videos on our project website. These findings highlight MuJoCo Playground’s capacity for training pixel-based policies that transfer reliably to real hardware in a zero-shot manner. Additional implementation details are described in \Cref{appendix:MadronaMJXBenchmark}.

\subsection{Training Throughput}

\begin{figure}[t]
    \centering
    \includegraphics[width=1\linewidth]{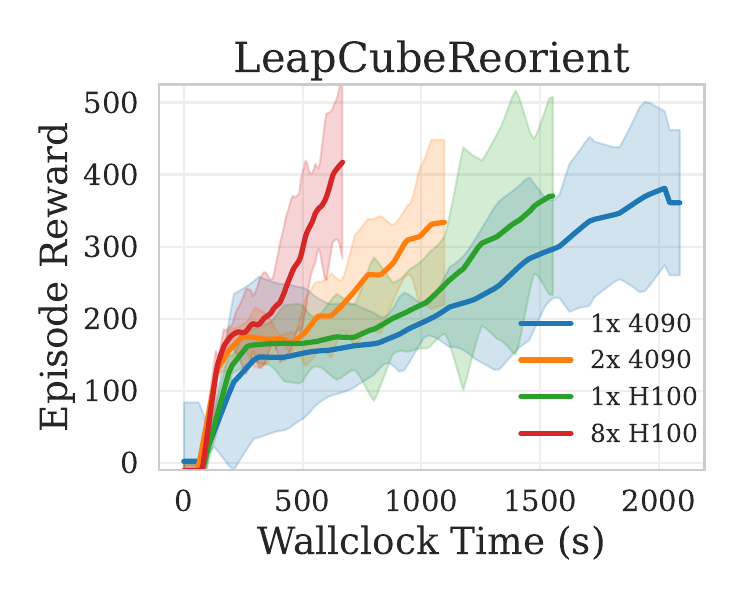}
    \caption{\small Training wallclock time for LeapCubeReorient on different GPU device topologies. 1x 4090 takes $\sim$ 2080 (s) to train and 8x H100 takes $\sim$ 670 (s) to train. All runs use the \emph{same hyperparams} (e.g. 8192 num envs); we leave tuning hyperparams per topology as a future exercise.}
    \label{fig:leap_cube_throughput}
\end{figure}

Across our sim-to-real studies, we used several GPU hardware setups and topologies, including NVIDIA RTX 4090, A100, and H100 GPUs. In \Cref{fig:leap_cube_throughput}, we break down the training performance of the LeapCubeReorient environment on different configurations for a fixed set of RL hyper-parameters, demonstrating that MJX is effective on both consumer-grade and datacenter graphics cards. We see that GPUs with higher theoretical performance and larger topologies can reduce training time by a factor of 3x on a contact-rich task like in-hand reorientation. We leave optimization of topology-specific hyper-parameters as future work (e.g. the number of environments should ideally increase for larger topologies to maximize throughput, as long as the RL algorithm can utilize the increase in data per epoch). In \Cref{tab:dm_control_env_training_throughput}, \Cref{tab:locomotion_training_throughput}, and \Cref{tab:manipulation_training_throughput} in the appendix, we report RL training throughput for all environments in MuJoCo Playground on a single~A100~GPU.

\subsubsection{Training Throughput with Batch Rendering}
\label{sec:madrona_bottlenecks_mini}

\begin{figure}[t]
    \centering
    \includegraphics[width=1\linewidth]{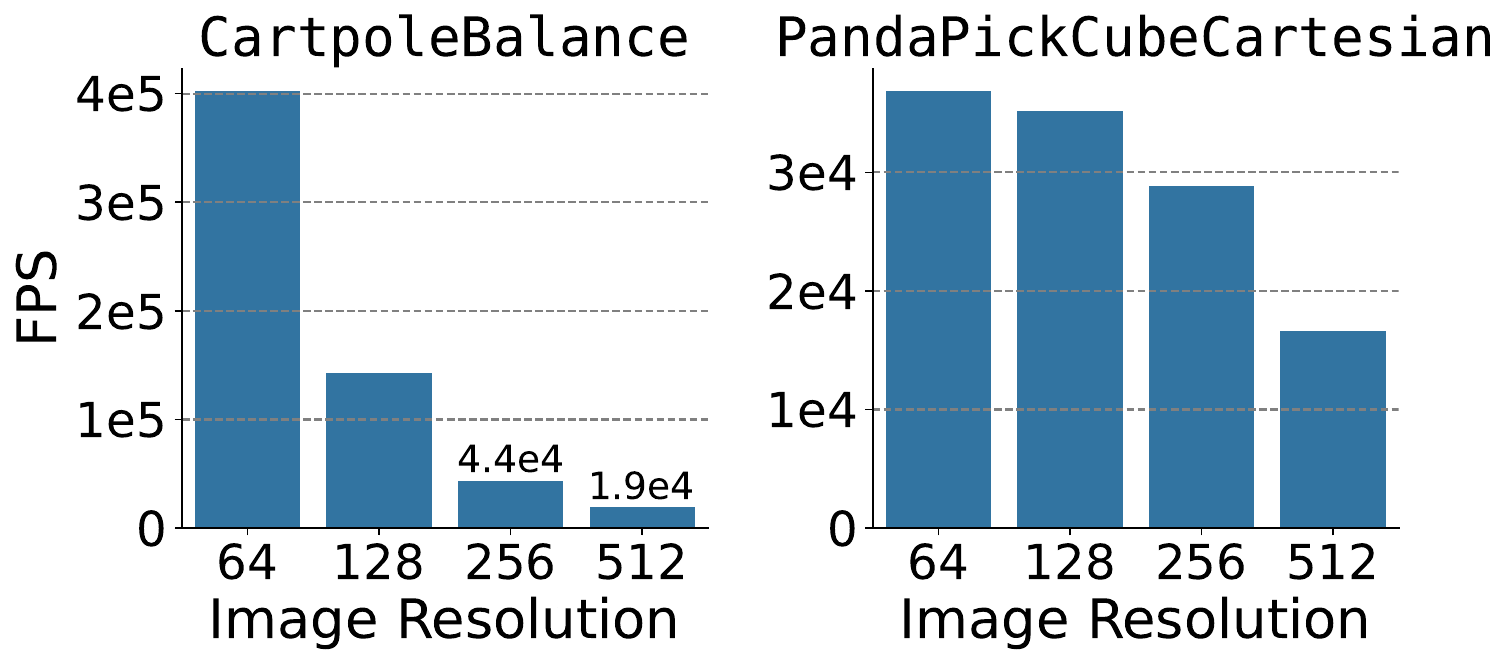}
    \caption{\small Environment steps per second on the single-camera CartpoleBalance and PandaPickCubeCartesian environments with pixel-based observations from our on-device renderer.}
    \label{fig:madrona_mjx_main}
\end{figure}

\Cref{fig:madrona_mjx_main} highlights the throughput of stepping two of our environments with pixel observations at different resolutions. By pairing MJX physics with Madrona batch rendering, our Cartpole and Franka environments unroll at roughly 403,000 and 37,000 steps per second respectively. Note that our Franka physics are over 20x more costly than Cartpole's, resulting in the lower sensitivity of FPS to image resolution.

Computationally, pixel-based policy training generally involves four main components: physics simulation, observation rendering, policy inference and policy updates. \Cref{fig:madrona_mjx_main} only encapsulates the former two and is not fully indicative of overall training throughput.

We find that in the context of a PPO training loop, physics, rendering, and inference together only comprise 9\% and 43\% of the Cartpole and Franka total training times, respectively, with most of the time spent updating the expensive CNN-based networks. Hence, compared to traditional on-policy training pipelines, we have shifted our bottleneck from collecting data to processing it. Training bottlenecks are further discussed in \Cref{sec:appendix_madrona_bottlenecks} under \Cref{tab:madrona_training_breakdown_measured} and \Cref{tab:madrona_training_breakdown_calc}. Further performance benchmarking and a rough comparison against prior simulators are in \Cref{sec:madrona_benchmarking}.

\subsubsection{RL Libraries}
\label{sec:results_different_rl_libs}

\begin{figure}[t]
    \centering
    \includegraphics[width=1\linewidth]{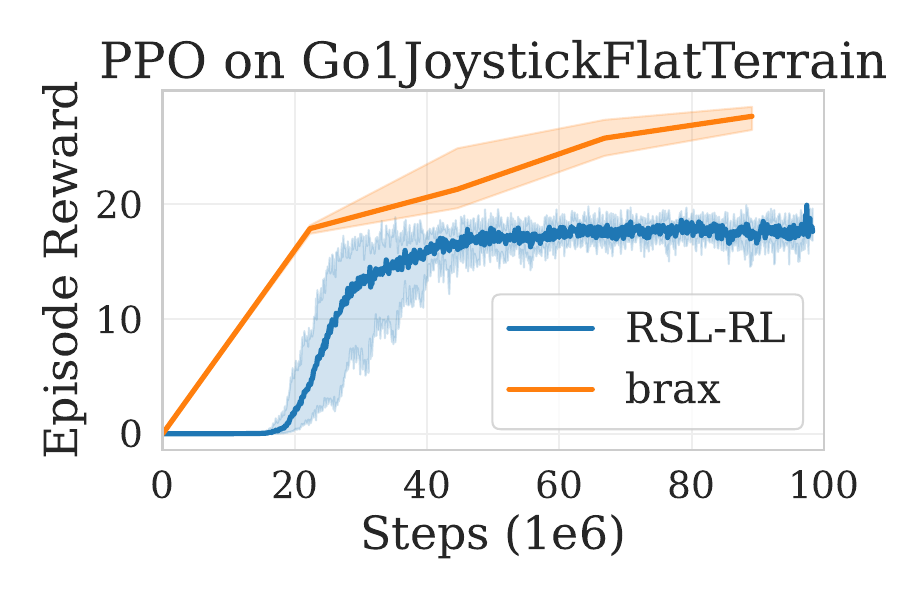}
    \caption{\small Reward curves for PPO trained with RSL-RL and brax on an RTX-4090 GPU for 3 seeds each on the Unitree Go1.}
    \label{fig:torch_comparison_go1}
\end{figure}

While MuJoCo Playground primarily uses a JAX-based physics simulator, practitioners are able to use both JAX and torch-based RL libraries for training RL agents. In \Cref{fig:torch_comparison_go1}, we show reward curves for PPO agents trained using both Brax~\cite{freeman2021brax} and RSL-RL \cite{rsl_rl} implementations. Each corresponding RL library is trained with custom hyperparameters tailored to the corresponding PPO implementation. Both libraries are able to achieve successful rewards and gaits within similar wallclock times. All other results in this paper were obtained using the Brax PPO and SAC implementations.

%% file: sections/5_relatedwork.tex
\section{Related Work}

\paragraph{Physics simulation on GPU} The PhysX GPU implementation \cite{liang2018gpu} has been heavily relied on for robotic sim-to-real workloads via IsaacGym \cite{makoviychuk2021isaac} and more recently Isaac Lab \cite{mittal2023orbit}. The PhysX GPU implementation, however, is closed-source~\cite{liang2018gpu} and researchers lack the ability to extend the simulator for their specific tasks or workloads. Several GPU-based physics engines are open-source, such as MJX \cite{todorov2012mujoco, mujocoxla}, Brax \cite{freeman2021brax}, Warp \cite{macklin2022warp}, and Taichi \cite{hu2019difftaichi}. Only a limited set of robot environments~\cite{sferrazza2024humanoidbench, xue2024full} leverage these open-source counterparts, in contrast to the wide range of robotic sim-to-real results that were achieved with IsaacGym and Isaac Lab. Most recently, Genesis \cite{Genesis} provides a rigid-body implementation similar to MJX implemented using Taichi, that allows for dynamic constraints/contacts. However, sim-to-real results are still limited to a few locomotion policies.

\paragraph{Sim-to-real RL} A variety of locomotion and manipulation policies have successfully been deployed in the real world zero-shot \cite{cheng2024extreme, zhuang2023robot, li2024reinforcement, long2024learning, radosavovic2024learning, singh2024dextrah, cho2024corn}. We complement these results by demonstrating zero-shot sim-to-real on the Leap Hand, Unitree Go1, Berkeley Humanoid, Unitree G1, Booster T1, and Franka arm using MuJoCo rather than closed-source simulators. Similar to \cite{mittal2023orbit, makoviychuk2021isaac}, we provide code for environments and training.

\paragraph{Vision-based RL} State-of-the-art algorithms such as DrQ \cite{yarats2021drqv2}, RL from Augmented Data (RAD) \cite{laskin_lee2020rad}, Dreamerv3 \cite{hafner2023dreamerv3}, TD-MPC2 \cite{hansen2024tdmpc2}, and EfficientZeroV2 \cite{wang2024efficientzero} have pushed pixel-based RL performance over the years. Transferring these advances to the real world is appealing, as visual control loops offer precise positioning and robust behaviour in uncontrolled in-the-wild scenarios \cite{haiderbhai2024cutrope}. The limitation of training directly from pixel data is the large visual sim-to-real gap between simulation and reality, which is often overcome using domain randomization \cite{domainrand2017}. However, such training methods require exponentially more training samples. As as result, policies are typically trained with proprioceptive observations in simulation and subsequently distilled into vision-based policies offline \cite{chen2022system, haarnoja2024learning, cheng2024extreme}, or trained with smaller exteroceptive observations \cite{miki2022learning, agarwal2023legged}. With Madrona, we are able to train vision-based policies directly in simulation without a distillation step using high-throughput batch rendering, similar to \cite{mittal2023orbit} and \cite{tao2024maniskill3}.

%% file: sections/8_limitations.tex
\section{Limitations}

MuJoCo Playground inherits the \href{https://mujoco.readthedocs.io/en/stable/mjx.html#mjx-the-sharp-bits}{limitations of MJX} due to constraints imposed by JAX. First, just-in-time (JIT) compilation can be slow (1-3 minutes on Playground's tasks). Second, computation time related to contacts does not scale like the number of \emph{active} contacts in the scene, but like the number of \emph{possible} contacts in the scene. This is due to JAX's requirement of static shapes at compile time. This limitation can be overcome by using more flexible frameworks like Warp~\cite{macklin2022warp} and Taichi~\cite{Genesis}. This upgrade is an active area of development. Finally we should note that the vision-based training using Madrona is still at an early stage.

%% file: sections/6_conclusion.tex
\section{Conclusion}

MuJoCo Playground is a library built upon the open-source MuJoCo simulator and Madrona batch renderer with implementations across several reinforcement learning and robotics environments. We demonstrate policy training on various GPU topologies using JAX and pytorch-based reinforcement learning libraries. We also demonstrate sim-to-real deployment on several robotic tasks and embodiments, from locomotion to both dexterous and non-prehensile manipulation from proprioceptive state and from pixels. We look forward to seeing the community put this resource to use in advancing robotics research and its applications.

%% file: sections/7_ack.tex
\section*{Acknowledgments} 

We thank Jimmy Wu, Kyle Stachowicz, Kenny Shaw, and Zhongyu Li for help with hardware. We thank Dongho Khang and Yunhao Cao for help with locomotion. We thank Rushrash Hari for help with hardware. We thank Luc Guy Rosenzweig, Brennan Shacklett and Kayvon Fatahalian for their extensive support in integrating the Madrona project into MJX. We thank Ankur Handa for help with manipulation. We thank Laura Smith and Philipp Wu for always being there to help with any problem and answer any question. We thank Lambda labs for sponsoring compute for the project. We thank Stone Tao for discussions on manipulation environments in MJX. We thank Erwin Coumans for introducing us to the Madrona team. We thank Kevin Bergamin and Michael Lutter for fruitful technical discussions and paper draft feedback. We thank Brent Yi for fruitful technical discussions and help with the website. We thank Lambda Labs for supporting this project with cloud compute credits.

This work is supported in part by The AI Institute. K. Sreenath has financial interest in Boston Dynamics AI Institute LLC.  He and the company may benefit from the commercialization of the results of this research.

This work was supported in part by the ONR Science of Autonomy Program N000142212121 and the BAIR Industrial Consortium. Pieter Abbeel holds concurrent appointments as a Professor at UC Berkeley and as an Amazon Scholar. This paper describes work performed at UC Berkeley and is not associated with Amazon.

%% file: appendix/dm_control.tex
\section{DM Control Suite}
\label{sec:appendix_dm_control}

\subsection{Environments}
\label{sec:appendix_dm_control_envs}

In \Cref{tab:dm_control_envs}, we show the environments from DM Control Suite (\cite{tassa2018deepmind}) that were re-implemented in MuJoCo Playground. Certain XMLs were modified for performance and are shown in the table. 

\begin{table*}[!ht]
\centering
\begin{tabular}{|l|p{1.2cm}|p{10cm}|} %
\hline
\textbf{Env} & \textbf{MJX} & \textbf{XML Modifications} \\ \hline
acrobot-swingup           & $\checkmark$ & iterations=2, ls\_iterations=4 \\ \hline
acrobot-swingup\_sparse    & $\checkmark$ &  \\ \hline
ball\_in\_cup-catch         & $\checkmark$ & iterations=1, ls\_iterations=4 \\ \hline
cartpole-balance          & $\checkmark$ & iterations=1, ls\_iterations=4 \\ \hline
cartpole-balance\_sparse   & $\checkmark$ &  \\ \hline
cartpole-swingup          & $\checkmark$ &  \\ \hline
cartpole-swingup\_sparse   & $\checkmark$ &  \\ \hline
cheetah-run               & $\checkmark$ & iterations=4, ls\_iterations=8, max\_contact\_points=6, max\_geom\_pairs=4 \\ \hline
finger-spin               & $\checkmark$ & iterations=2, ls\_iterations=8, max\_contact\_points=4, max\_geom\_pairs=2, removed cylinder collision \\ \hline
finger\_turn\_easy          & $\checkmark$ &  \\ \hline
finger\_turn\_hard          & $\checkmark$ &  \\ \hline
fish-upright              & $\checkmark$ & iterations=2, ls\_iterations=6, disabled contacts \\ \hline
fish-swim                 & $\checkmark$ &  \\ \hline
hopper-stand              & $\checkmark$ & iterations=4, ls\_iterations=8, max\_contact\_points=6, max\_geom\_pairs=2 \\ \hline
hopper-hop                & $\checkmark$ &  \\ \hline
humanoid-stand            & $\checkmark$ & timestep=0.005, max\_contact\_points=8, max\_geom\_pairs=8 \\ \hline
humanoid-walk             & $\checkmark$ &  \\ \hline
humanoid-run              & $\checkmark$ &  \\ \hline
pendulum-swingup          & $\checkmark$ & timestep=0.01, iterations=4, ls\_iterations=8 \\ \hline
point\_mass-easy           & $\checkmark$ & iterations=1, ls\_iterations=4 \\ \hline
reacher-easy              & $\checkmark$ & timestep=0.005, iterations=1, ls\_iterations=6 \\ \hline
reacher-hard              & $\checkmark$ &  \\ \hline
swimmer-swimmer6          & $\checkmark$ & timestep=0.003, iterations=4, ls\_iterations=8, contype/conaffinity set to 0 \\ \hline
swimmer-swimmer15         & $\times$ &  \\ \hline
walker-stand              & $\checkmark$ & timestep=0.005, iterations=2, ls\_iterations=5, max\_contact\_points=4, max\_geom\_pairs=4 \\ \hline
walker-walk               & $\checkmark$ &  \\ \hline
walker-run                & $\checkmark$ &  \\ \hline
manipulator-bring\_ball    & $\times$ & \\ \hline
manipulator-bring\_peg     & $\times$ &  \\ \hline
manipulator-insert\_ball   & $\times$ &  \\ \hline
manipulator-insert\_peg    & $\times$ &  \\ \hline
dog-stand                 & $\times$ &  \\ \hline
dog-walk                  & $\times$ &  \\ \hline
dog-trot                  & $\times$ &  \\ \hline
dog-run                   & $\times$ &  \\ \hline
dog-fetch                 & $\times$ &  \\ \hline
\hline
\end{tabular}
\caption{DM Control Suite Environments ported to MJX. Where specified, XML modifications were made to the solver iterations, line search iterations, as well as contact custom parameters for MJX.}
\label{tab:dm_control_envs}
\end{table*}

\subsection{RL Training Results}
\label{sec:appendix_dm_control_curves}

\begin{figure}[ht]
    \centering
    \includegraphics[width=1.0\linewidth]{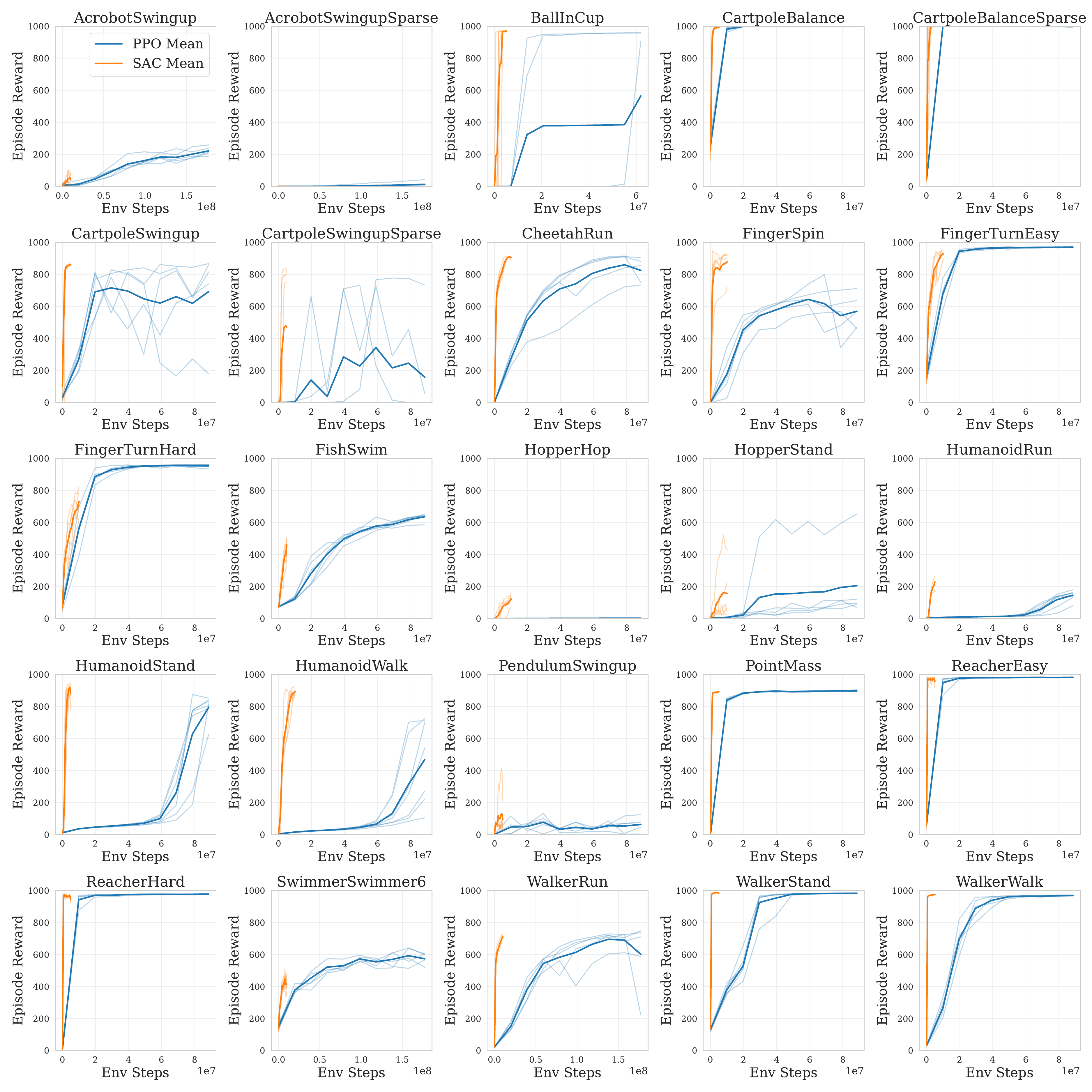}
    \caption{\small Reward vs environment steps for PPO and SAC on the full DM Control Suite environments in MuJoCo Playground. We run PPO for 60M steps, with a few selected environments running on 100M steps. SAC runs for 5M steps. All settings are run with 5 seeds on a single A100 GPU device.}
    \label{fig:dm_control_step_reward}
\end{figure}

\begin{figure}[ht]
    \centering
    \includegraphics[width=1.0\linewidth]{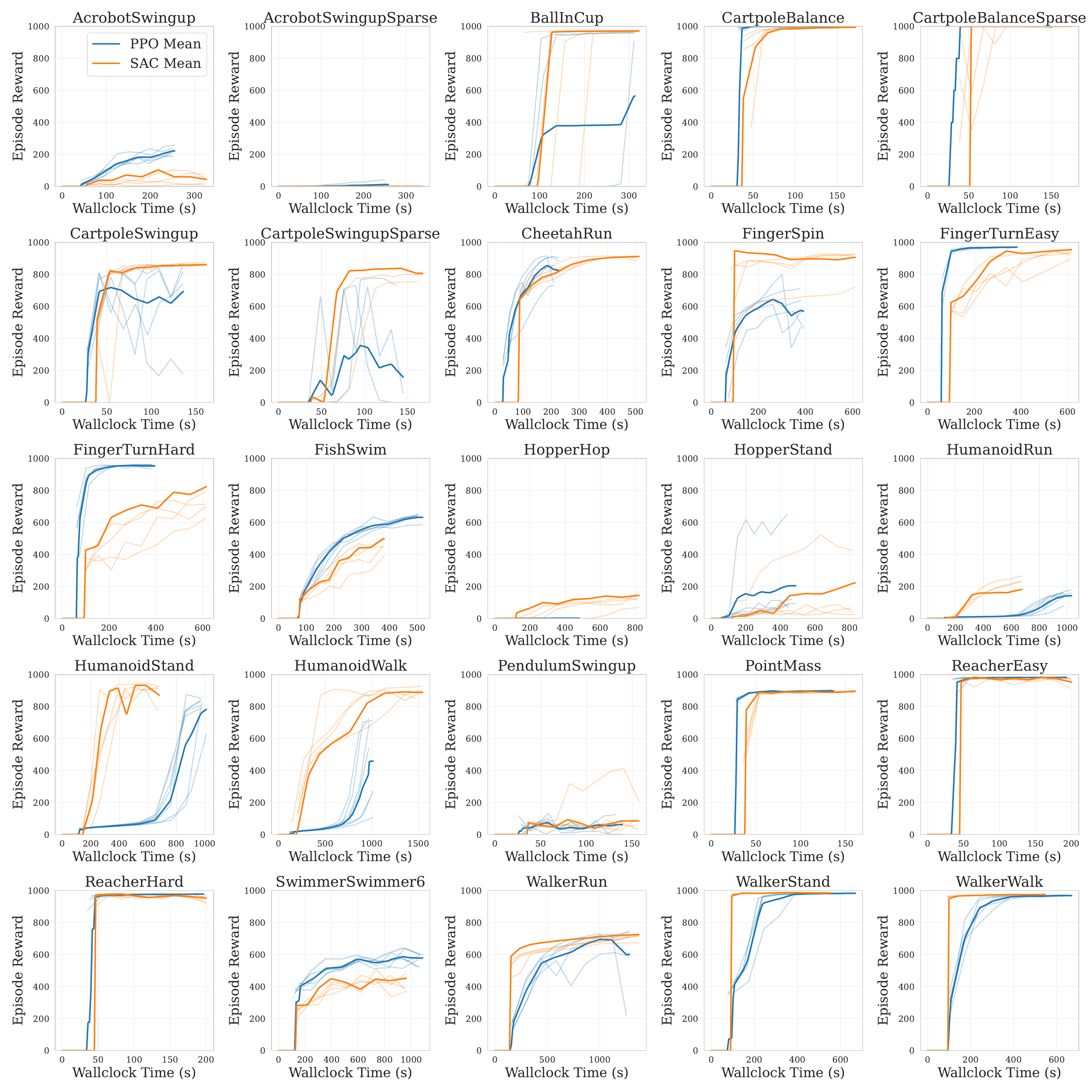}
    \caption{\small Reward vs wallclock time for PPO and SAC on the full DM Control Suite environments in MuJoCo Playground. All settings are run with 5 seeds on a single A100 GPU device.}
    \label{fig:dm_control_time_reward}
\end{figure}

For all DM Control Suite environments ported to MuJoCo Playground, we train both PPO~\cite{schulman2017proximal} and SAC~\cite{haarnoja2018soft} using the RL implementations in \cite{freeman2021brax} and we report reward curves below. In \Cref{fig:dm_control_step_reward} we report environment steps versus reward and in \Cref{fig:dm_control_time_reward} we report wallclock time versus reward. All environments are run across 5 seeds on a single A100 GPU.

\subsection{RL Training Throughput}
\label{sec:appendix_dm_control_training_throughput}

We report training throughput on all DM Control Suite environments in \Cref{tab:dm_control_env_training_throughput} by dividing the number of environment steps by wallclock time, as reported in \Cref{sec:appendix_dm_control_curves}, for each RL algorithm.

\begin{table*}[ht]
\centering
\begin{tabular}{|l|c|c|}
\hline 
 Env                   & PPO Steps per Second     & SAC Steps Per Second     \\
\hline 
 AcrobotSwingup        & 752092 $\pm$ 11562 & 30661 $\pm$ 244  \\
 AcrobotSwingupSparse  & 750597 $\pm$ 4640  & 30624 $\pm$ 210  \\
 BallInCup             & 235899 $\pm$ 565   & 15492 $\pm$ 283  \\
 CartpoleBalance       & 718626 $\pm$ 6894  & 30891 $\pm$ 168  \\
 CartpoleBalanceSparse & 721061 $\pm$ 14135 & 31031 $\pm$ 183  \\
 CartpoleSwingup       & 728088 $\pm$ 12503 & 30870 $\pm$ 207  \\
 CartpoleSwingupSparse & 718355 $\pm$ 10189 & 31061 $\pm$ 226  \\
 CheetahRun            & 435162 $\pm$ 12183 & 18819 $\pm$ 202  \\
 FingerSpin            & 246791 $\pm$ 1763  & 16475 $\pm$ 153  \\
 FingerTurnEasy        & 245255 $\pm$ 4561  & 16086 $\pm$ 112  \\
 FingerTurnHard        & 245421 $\pm$ 4278  & 16084 $\pm$ 69   \\
 FishSwim              & 183750 $\pm$ 1773  & 11591 $\pm$ 55   \\
 HopperHop             & 201313 $\pm$ 2833  & 12098 $\pm$ 166  \\
 HopperStand           & 201517 $\pm$ 3227  & 12008 $\pm$ 255  \\
 HumanoidRun           & 91617 $\pm$ 1019   & 5886 $\pm$ 62    \\
 HumanoidStand         & 91927 $\pm$ 1004   & 5893 $\pm$ 17    \\
 HumanoidWalk          & 91563 $\pm$ 1150   & 5842 $\pm$ 51    \\
 PendulumSwingup       & 724126 $\pm$ 21524 & 32836 $\pm$ 178  \\
 PointMass             & 730775 $\pm$ 3608  & 31710 $\pm$ 148  \\
 ReacherEasy           & 520021 $\pm$ 9637  & 24888 $\pm$ 149  \\
 ReacherHard           & 523441 $\pm$ 8012  & 24874 $\pm$ 156  \\
 SwimmerSwimmer6       & 167259 $\pm$ 2377  & 10012 $\pm$ 79   \\
 WalkerRun             & 141581 $\pm$ 831   & 6069 $\pm$ 48    \\
 WalkerStand           & 140360 $\pm$ 1762  & 6085 $\pm$ 29    \\
 WalkerWalk            & 139818 $\pm$ 1267  & 6098 $\pm$ 30    \\
\hline 
\end{tabular}
\caption{Training throughput is displayed for all the DM Control Suite environments on an A100 GPU device across 5 seeds using brax PPO and the RL hyperparameters in Appendix \Cref{sec:rl_hypers}. We report the 95th percentile confidence interval.}
\label{tab:dm_control_env_training_throughput}
\end{table*}

\clearpage

%% file: appendix/locomotion.tex
\section{Locomotion}
\label{sec:appendix_locomotion}

\subsection{Environment}
\label{sec:appendix_locomotion_env}

In \Cref{tab:locomotion_envs} we show all the locomotion environments available in MuJoCo Playground, broken down by robot platform and available controller.

\begin{table*}[!ht]
\centering
\begin{tabular}{|l|l|p{10cm}|} %
\hline
\textbf{Robot} & \textbf{Type} & \textbf{Environment} \\ \hline
Google Barkour     & Quadruped      & JoystickFlatTerrain, JoystickRoughTerrain \\ \hline
Berkeley Humanoid  & Biped         & Joystick \\ \hline
Unitree G1      & Biped     & Joystick \\ \hline
Booster T1      & Biped     & Joystick \\ \hline
Unitree Go1      & Quadruped     & JoystickFlatTerrain, JoystickRoughTerrain, Getup, \mbox{Handstand,} Footstand \\ \hline
Unitree H1 & Biped & InplaceGaitTracking, JoystickGaitTracking  \\ \hline
OP3 & Biped & Joystick \\ \hline
Boston Dynamics Spot & Quadruped & JoystickFlatTerrain, JoystickGaitTracking, Getup \\ \hline
\hline
\end{tabular}
\caption{Locomotion environments implemented in MuJoCo Playground by robot platform.}
\label{tab:locomotion_envs}
\end{table*}

\subsection{RL Training Details}

\subsubsection{Observation and Action}
We use a unified observation space across all locomotion environments:
\begin{enumerate}[label=(\alph*)]
    \item Gravity projected in the body frame,
    \item Base linear and angular velocity,
    \item Joint positions and velocities,
    \item Previous action,
    \item (Optional) User command for joystick-based tasks.
\end{enumerate}

For humanoid locomotion tasks, a phase variable~\cite{shao2021learning} is introduced to shape the gait. This phase variable cycles between $-\pi$ and $\pi$ for each foot, representing the gait phase. To capture this information effectively, the $\cos$ and $\sin$ of the phase variable for each foot are included in the observation space. This representation provides a continuous and smooth encoding of the phase, enabling the policy to synchronize its actions with the desired gait cycle.

The action space is defined differently depending on the task. 
For joystick tasks, we use an \emph{absolute} joint position with a default offset:
\[
    q_{\text{des}, t} = q_{\text{default}} + k_a \, a_t,
\]
where \(k_a\) is the action scale. 
For all other tasks, we use a \emph{relative} joint position:
\[
    q_{\text{des}, t} = q_{\text{des}, t-1} + k_a \, a_t.
\]
The desired joint position is mapped to torque via a PD controller:
\begin{equation}
    \tau = k_p \bigl(q_{\text{des}} - q\bigr) \;-\; k_d \,\dot{q},
    \label{eq:pd_mapping}
\end{equation}
where \(k_p\) and \(k_d\) are the proportional and derivative gains, respectively.

\subsubsection{Domain Randomization}
To reduce the sim-to-real gap, we randomize several parameters during training:
\begin{itemize}
    \item \textbf{Sensor noise:} All sensor readings are corrupted with noise.
    \item \textbf{Dynamic properties:} Physical parameters that are difficult to measure precisely 
    (e.g., link center-of-mass, reflected inertia, joint calibration offsets).
    \item \textbf{Task uncertainties:} Ground friction and payload mass.
\end{itemize}

\subsubsection{Reward and Termination}

\begin{table}[h!]
\centering
\caption{Reward Functions}
\label{table:reward-functions}
\begin{tabular}{|l|l|}
\hline
\textbf{Reward}                  & \textbf{Expression}                                                                                         \\ \hline
Linear Velocity Tracking         & $r_v = k_v \exp\left(-\|cmd_{v,xy} - v_{xy}\|^2 / \sigma_v\right)$                                          \\ \hline
Angular Velocity Tracking        & $r_\omega = k_\omega \exp\left(-\|cmd_{\omega,z} - \omega_z\|^2 / \sigma_\omega\right)$                     \\ \hline
Feet Airtime                     & $r_\text{air} = \text{clip}\left((T_\text{air} - T_\text{min}) \cdot C_\text{contact}, 0, T_\text{max} - T_\text{min}\right)$ \\ \hline
Feet Clearance                   & $r_\text{clear} = k_\text{clear} \cdot \|p_{f,z} - p^\text{des}_{f,z}\|^2 \cdot \|v_{f,xy}\|^{0.5}$          \\ \hline
Feet Phase                       & $r_\text{phase} = k_\text{phase} \cdot \exp\left(-\|p_{f,z} - r_z(\phi)\|^2 / \sigma_\text{phase}\right)$    \\ \hline
Feet Slip                        & $r_\text{slip} = k_\text{slip} \cdot \|C_{f,i} \cdot v_{f,xy}\|^2$                                          \\ \hline
Orientation                      & $r_\text{ori} = k_\text{ori} \cdot \| \phi_\text{body,xy}\|^2$                                              \\ \hline
Joint Torque                     & $r_\tau = k_\tau \cdot \|\tau\|^2$                                                                         \\ \hline
Joint Position                   & $r_q = k_q \cdot \|q - q_\text{nominal}\|^2$                                                               \\ \hline
Action Rate                      & $r_\text{rate} = k_\text{rate} \cdot \|a_t - a_{t-1}\|^2$                                                  \\ \hline
Energy Consumption               & $r_\text{energy} = k_\text{energy} \cdot \|\dot{q} \cdot \tau\|$                                           \\ \hline
Pose Deviation                   & $r_\text{pose} = k_\text{pose} \cdot \exp\left(-\|q - q_\text{default}\|^2\right)$                         \\ \hline
Termination (Penalty)            & $r_\text{termination} = k_\text{termination} \cdot \text{done}$                                            \\ \hline
Stand Still (Penalty)            & $r_\text{standstill} = k_\text{standstill} \cdot \|cmd_{v,xy}\|$                                           \\ \hline
Linear Velocity in Z (Penalty)   & $r_\text{lin\_z} = k_\text{lin\_z} \cdot \|v_{z}\|^2$                                                      \\ \hline
Angular Velocity in XY (Penalty) & $r_\text{ang\_xy} = k_\text{ang\_xy} \cdot \|\omega_{x,y}\|^2$                                             \\ \hline
\end{tabular}
\end{table}

In Table~\ref{table:reward-functions}, $cmd_{v,xy}$ and $cmd_{\omega,z}$ represent the commanded linear velocity in the $xy$-plane and angular velocity around the $z$-axis, respectively. $v_{xy}$ and $\omega_z$ are the actual linear and angular velocities. $T_s$ and $T_a$ represent the time of the last touchdown and takeoff of the feet. $p_{f,z}$ and $p^\text{des}_{f,z}$ denote the actual and desired foot heights, while $v_{f,xy}$ is the horizontal foot velocity. $\tau$ is the torque, $q$ is the joint position, and $\dot{q}$ is the joint velocity.

The total reward $r_\text{total}$ is calculated as the weighted sum of all the reward terms:
\[
r_\text{total} = \sum_i w_i r_i,
\]
Finally, the total reward is clipped to ensure it remains non-negative.

\textbf{Termination:} For joystick-controlled policies, we use a reduced collision model (only the feet) and terminate the episode if the robot inverts (e.g., ends up upside down). For other tasks, we employ the full collision model approximated using geometric primitives.

\subsubsection{Network Architecture}

We employ an asymmetric actor--critic~\cite{pinto2018asymmetric} setup, in which the policy network (actor) and the value network (critic) receive different observation inputs. The policy network is fed with the aforementioned observations, while the value network additionally receives uncorrupted versions of these signals and extra sensor readings such as contact forces, perturbation forces, and joint torques.

Both the policy and value networks use a three-layer multilayer perceptron (MLP) with hidden sizes of 512, 256, and 128. Each hidden layer uses the Swish \cite{ramachandran2017searchingactivationfunctions} activation function. A full set of hyper-parameters is available in \Cref{sec:rl_hypers}.

\subsubsection{Finetuning}
\paragraph{Joystick policy}
\begin{enumerate}
    \item Train for 100\,M timesteps with a command range of \(\{1.5, 0.8, 1.2\}\).
    \item Finetune for 50\,M timesteps with a command range of \(\{1.5, 0.8, 2\pi\}\).
    \item Finetune on rough terrain for 100\,M timesteps.
\end{enumerate}

\paragraph{Getup policy}
\begin{enumerate}
    \item Train with a power termination cutoff of 400\,W.
    \item Finetune with a joint velocity cost.
\end{enumerate}

\paragraph{Handstand and footstand policies}
\begin{enumerate}
    \item Finetune with a joint acceleration and energy cost.
    \item Progressively reduce the power termination budget from 400\,W to 200\,W.
\end{enumerate}

Finally, all policies are trained on flat terrain for 200\,M timesteps, then finetuned on rough terrain for 100\,M timesteps. The rough terrain is modeled as a heightfield generated from Perlin noise.

\subsection{RL Training Results}
\label{sec:appendix_locomotion_curves}

\begin{figure}[ht]
    \centering
    \includegraphics[width=1.0\linewidth]{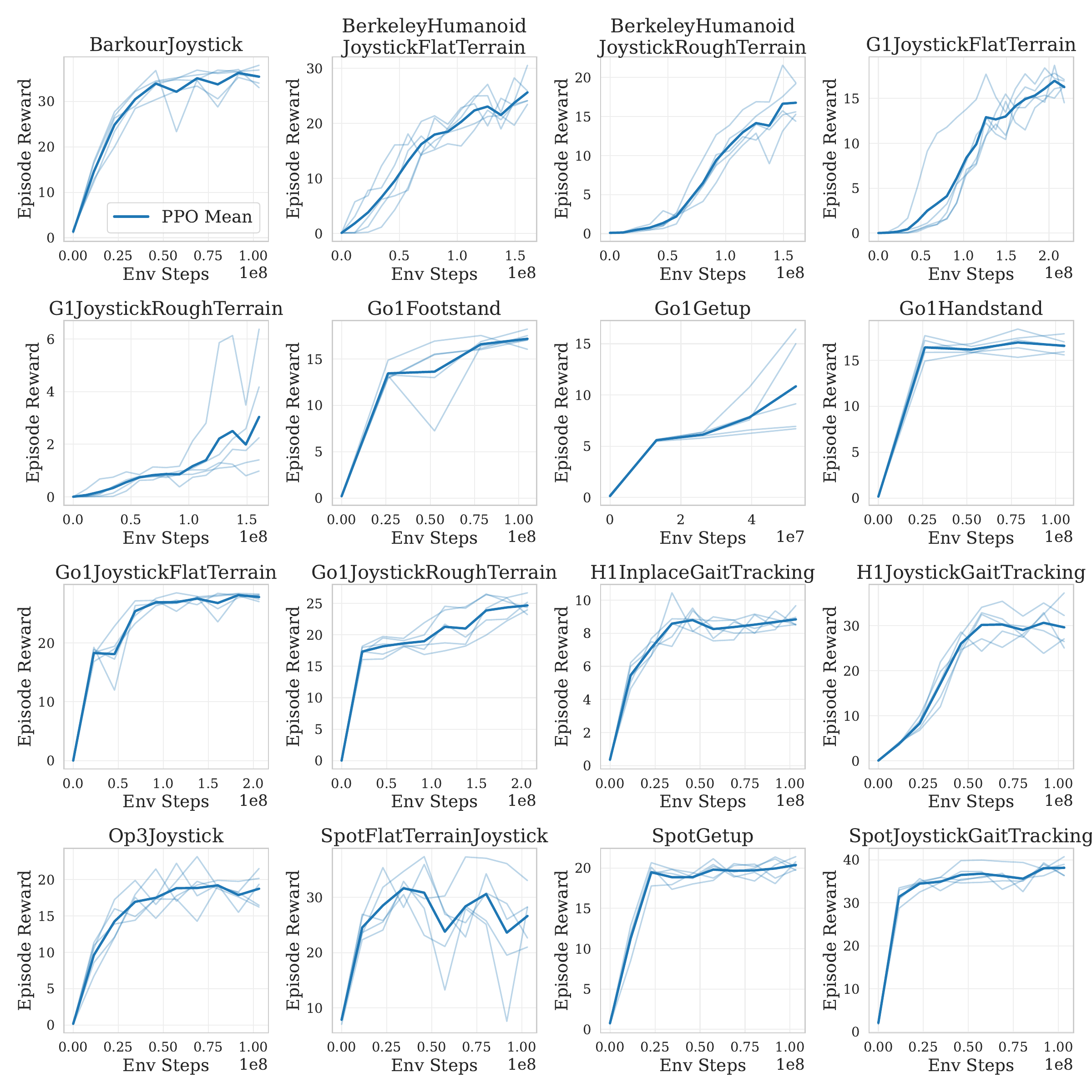}
    \caption{\small Reward vs environment steps for Brax PPO. All settings are run with 5 seeds on a single A100 GPU device.}
    \label{fig:locomotion_step_reward}
\end{figure}

\begin{figure}[ht]
    \centering
    \includegraphics[width=1.0\linewidth]{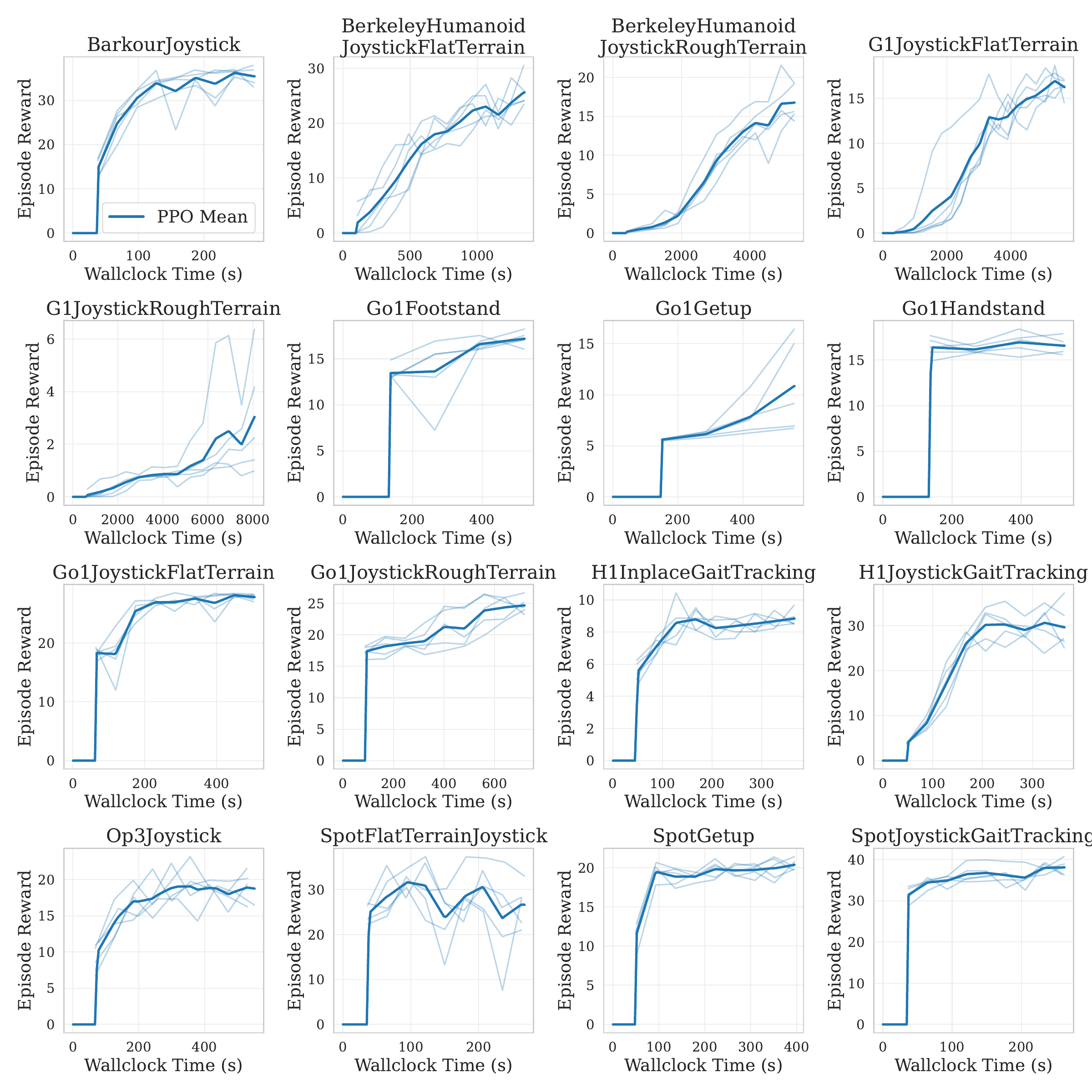}
    \caption{\small Reward vs wallclock time for Brax PPO. All settings are run with 5 seeds on a single A100 GPU device. Notice that the initial flat region measures the compilation time for the training + environment code.}
    \label{fig:locomotion_time_reward}
\end{figure}

For all locomotion environments implemented in MuJoCo Playground, we train with PPO using the RL implementation from \cite{freeman2021brax} and we report reward curves below. In \Cref{fig:locomotion_step_reward} we report environment steps versus reward and in \Cref{fig:locomotion_time_reward} we report wallclock time versus reward. All environments are run across 5 seeds on a single A100 GPU.

\subsection{RL Training Throughput}
\label{sec:appendix_locomotion_throughput}

\begin{figure}[t]
    \centering
    \includegraphics[width=0.8\linewidth]{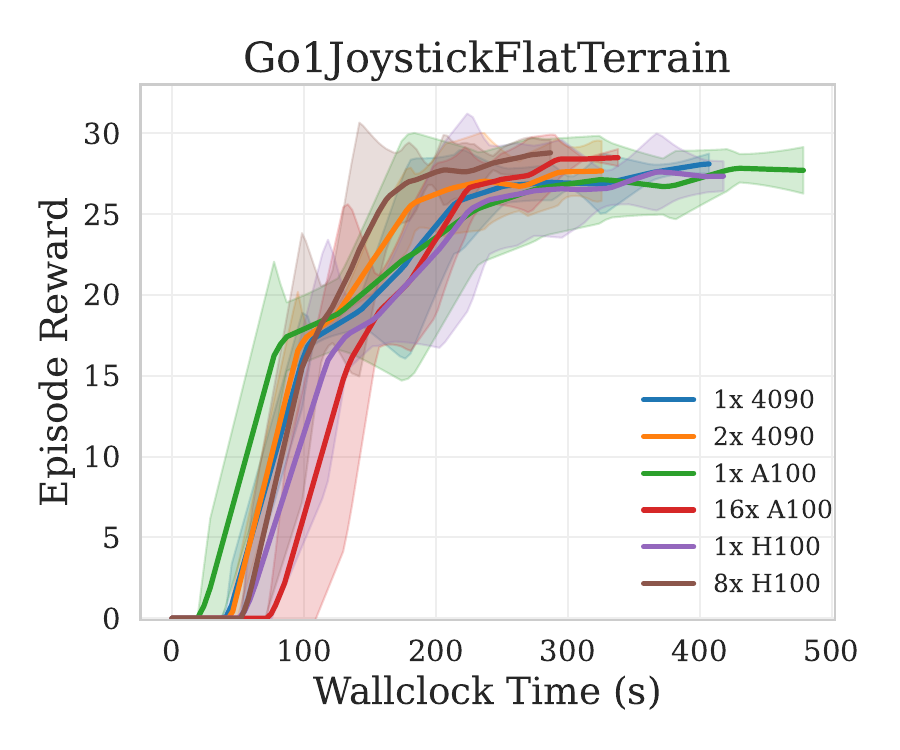}
    \caption{\small Training wallclock time for Go1JoystickFlatTerrain on different GPU devices and topologies.}
    \label{fig:go1_device_topo}
\end{figure}

In \Cref{tab:locomotion_training_throughput} we show training throughput for all locomotion envs. In \Cref{fig:go1_device_topo} we show training throughput of the Go1JoystickFlatTerrain environment. Different devices and topologies do not make material difference in training wallclock time, since the environment is quite simple with limited contacts between the feet and the floor.

\begin{table*}[ht]
\centering
\begin{tabular}{|l|c|c|}
\hline 
 Env                   & PPO Steps per Second     \\
\hline 
 BarkourJoystick                      & 385920 $\pm$ 2162 \\
 BerkeleyHumanoidJoystickFlatTerrain  & 120145 $\pm$ 484  \\
 BerkeleyHumanoidJoystickRoughTerrain & 30393 $\pm$ 44    \\
 G1Joystick                           & 106093 $\pm$ 131  \\
 Go1Footstand                         & 204578 $\pm$ 906  \\
 Go1Getup                             & 96173 $\pm$ 230   \\
 Go1Handstand                         & 204416 $\pm$ 738  \\
 Go1JoystickFlatTerrain               & 417451 $\pm$ 2955 \\
 Go1JoystickRoughTerrain              & 291060 $\pm$ 727  \\
 H1InplaceGaitTracking                & 289372 $\pm$ 1498 \\
 H1JoystickGaitTracking               & 291018 $\pm$ 1111 \\
 Op3Joystick                          & 198910 $\pm$ 406  \\
 SpotFlatTerrainJoystick              & 404931 $\pm$ 2710 \\
 SpotGetup                            & 266792 $\pm$ 1038 \\
 SpotJoystickGaitTracking             & 407572 $\pm$ 4091 \\
\hline 
\end{tabular}
\caption{Training throughput is displayed for all the Locomotion environments on an A100 GPU device across 5 seeds using brax PPO and the RL hyperparameters in \Cref{sec:rl_hypers}. We report the 95th percentile confidence interval.}
\label{tab:locomotion_training_throughput}
\end{table*}

\clearpage

\subsection{Real-world Setup}

All locomotion deployments are based on \texttt{ros2-control} and are written in \texttt{C++} with real-time guarantees. The Unitree SDK1, Unitree SDK2, and the Berkeley Humanoid EtherCAT master are each wrapped as abstract sensor and actuator hardware interfaces. These same interfaces are also used in Gazebo \cite{koenig2004design} to facilitate sim-to-sim verification.

Different RL policies can be loaded and executed within the same process—whether operating on physical hardware or in simulation—by receiving sensor readings and issuing control commands via the hardware interface. Each policy model is inferenced at 50 Hz using ONNX Runtime \cite{onnxruntime}, alongside a model-based estimator. In addition, a separate model-based estimator~\cite{Flayols_2017_estimator} runs at the hardware interface’s maximal communication frequency (500–2000 Hz), providing linear velocity observations and other diagnostic information.

\clearpage

%% file: appendix/manipulation.tex
\section{Manipulation}
\label{sec:appendix_manipulation}
\subsection{Environments}
\label{sec:appendix_manipulation_envs}

\begin{table*}[!ht]
\centering
\begin{tabular}{|l|p{10cm}|} %
\hline
\textbf{Robot}  & \textbf{Environment} \\ \hline
Aloha          & SinglePegInsertion \\ \hline
Franka Emika Panda &  PickCube, PickCubeOrientation, PickCubeCartesian, \mbox{OpenCabinet}  \\ \hline
Franka Emika Panda, Robotiq Gripper &  PushCube \\ \hline
Leap Hand & Reorient, RotateZAxis \\  \hline
\hline
\end{tabular}
\caption{Manipulation environments implemented in MuJoCo Playground by robot platform.}
\label{tab:manipulation_envs}
\end{table*}

\subsection{RL Training Results}
\label{sec:appendix_manipulation_curves}

\begin{figure}[ht]
    \centering
    \includegraphics[width=1.0\linewidth]{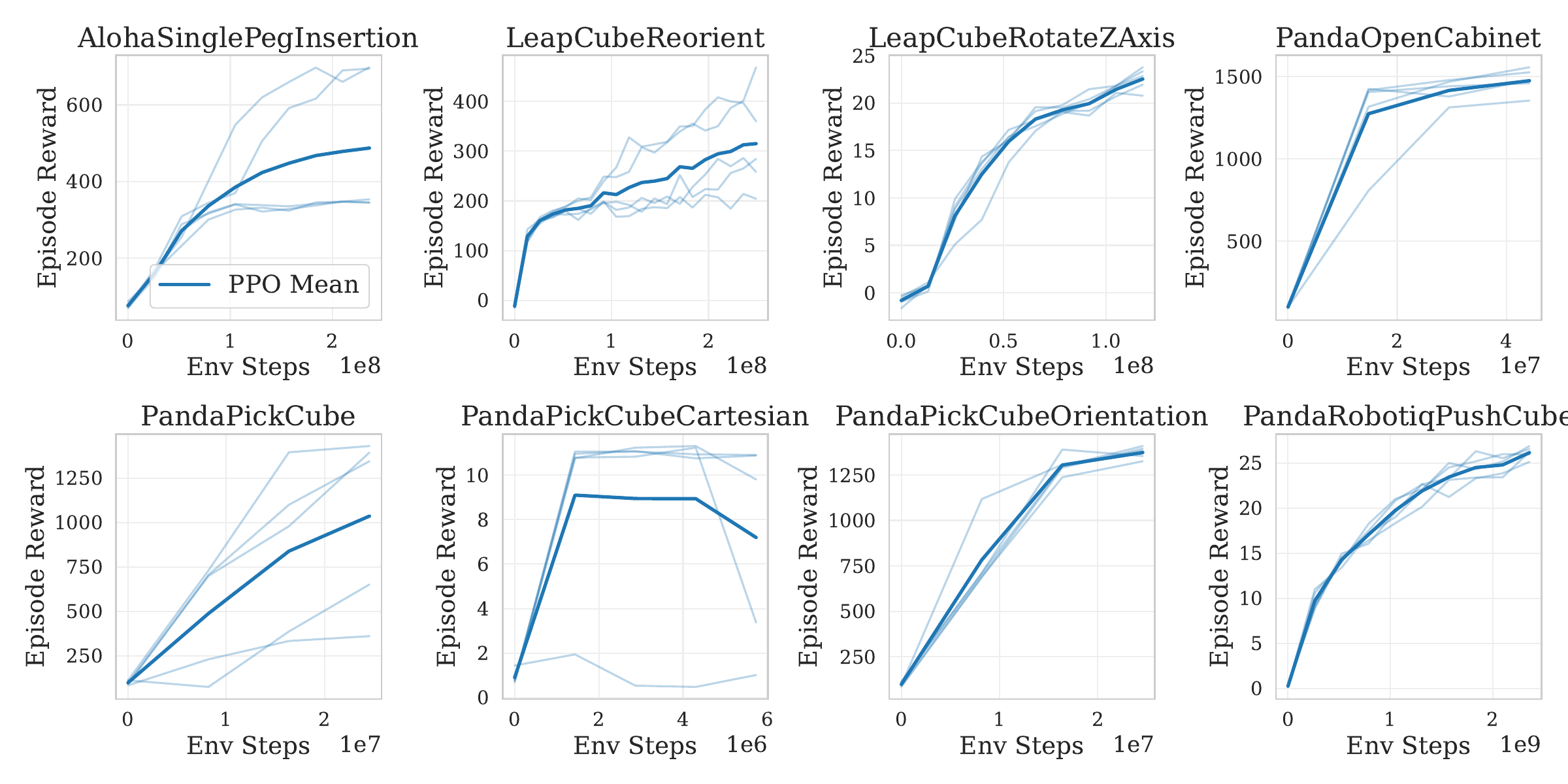}
    \caption{\small Reward vs environment steps for brax PPO. All settings are run with 5 seeds on a single A100 GPU device.}
    \label{fig:manipulation_step_reward}
\end{figure}

\begin{figure}[ht]
    \centering
    \includegraphics[width=1.0\linewidth]{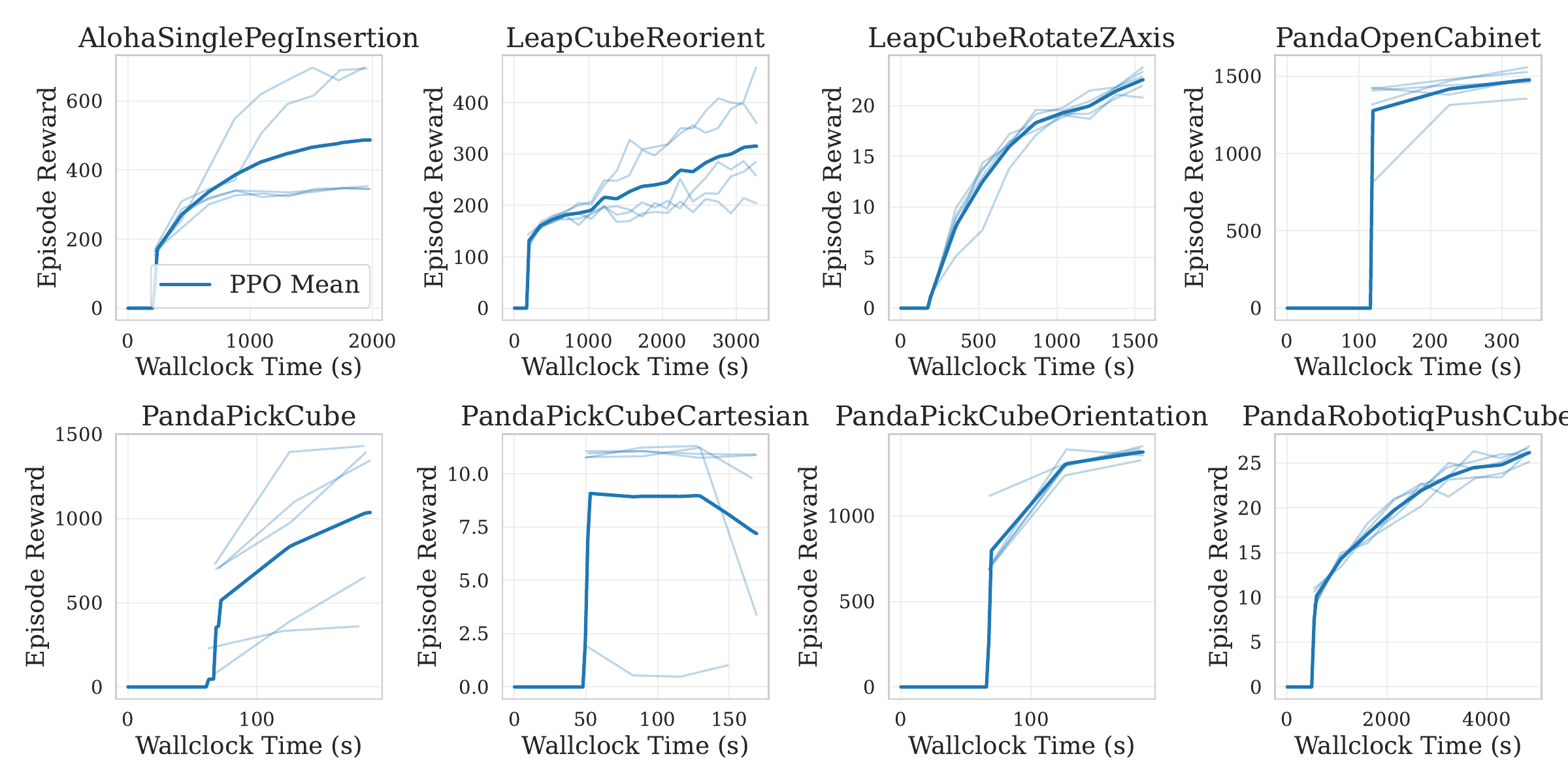}
    \caption{\small Reward vs wallclock time for brax PPO. All settings are run with 5 seeds on a single A100 GPU device. Notice that the initial flat region measures the compilation time for the training + environment code.}
    \label{fig:manipulation_time_reward}
\end{figure}

For all manipulation environments implemented in MuJoCo Playground, we train with PPO using the RL implementation from \cite{freeman2021brax} and we report reward curves below. In \Cref{fig:manipulation_step_reward} we report environment steps versus reward and in \Cref{fig:manipulation_time_reward} we report wallclock time versus reward. All environments are run across 5 seeds on a single A100 GPU.

\subsection{RL Training Throughput}
\label{sec:appendix_manipulation_throughput}

\begin{figure}[t]
    \centering
    \includegraphics[width=0.8\linewidth]{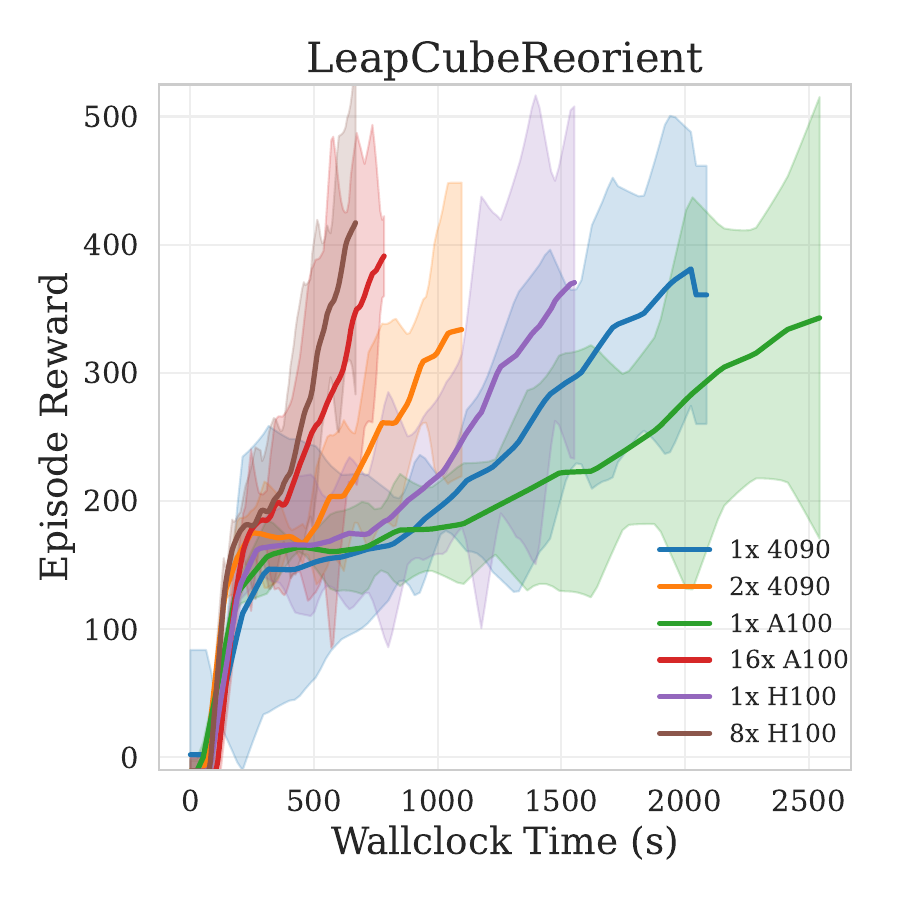}
    \caption{\small Training wallclock time for LeapHandReorient on different GPU devices and topologies.}
    \label{fig:leaphand_device_topo}
\end{figure}

We show RL training throughput for all manipulation environments below in \Cref{tab:manipulation_training_throughput}. In \Cref{fig:leaphand_device_topo} we show reward versus wallclock time on different GPU devices and topologies for the LeapCubeReorient environment.

\begin{table*}[ht]
\centering
\begin{tabular}{|l|c|c|}
\hline 
 Env                   & PPO Steps per Second     \\
\hline 
 AlohaSinglePegInsertion  & 121119 $\pm$ 2159 \\
 LeapCubeReorient         & 76354 $\pm$ 143   \\
 LeapCubeRotateZAxis      & 76602 $\pm$ 179   \\
 PandaOpenCabinet         & 136007 $\pm$ 1553 \\
 PandaPickCube            & 140386 $\pm$ 1707 \\
 PandaPickCubeCartesian   & 38015 $\pm$ 5302  \\
 PandaPickCubeOrientation & 140429 $\pm$ 1604 \\
 PandaRobotiqPushCube     & 487341 $\pm$ 4346 \\
\hline 
\end{tabular}
\caption{Training throughput is displayed for all the Manipulation environments on an A100 GPU device across 5 seeds using brax PPO and the RL hyperparameters in \Cref{sec:rl_hypers}. We report the 95th percentile confidence interval.}
\label{tab:manipulation_training_throughput}
\end{table*}

\clearpage

\subsection{Real-world Cube Reorientation with a Leap Hand}
\label{sec:appendix_leaphand_real}

In this section, we present the technical details of our real-world cube reorientation task using the LEAP Hand, covering the simulation environment, training process, hardware interface, and camera-based object pose estimation.

\subsubsection{Simulation Environment}

The in-hand reorientation environment is designed to sequentially re-orient a cube within the palm of a robotic hand, without dropping the cube. The cube is initialized randomly above the palm of the hand. The policy then receives the joint angle measurements, estimated cube pose, and previous action. Upon reaching a target orientation within a 0.4\,rad tolerance, a new orientation is sampled and the success counter is incremented. We continue re-sampling new target orientations until the cube is dropped or the hand becomes stuck for over 30\,s. To avoid trivial adjustments, new orientations are sampled at least 90° away from the previous goal (as in \cite{handa2023dextreme, li2024_drop}). The cube must reach a target orientation within 0.1\,rad (as opposed to 0.4\,rad in the real-world setup).

\paragraph{Policy Inputs and Actions}
As in locomotion environments, we use an asymmetric actor--critic setup, in which the policy network (actor) and the value network (critic) receive different observation inputs. The policy network is fed observations as outlined below, while the value network additionally receives uncorrupted robot pose, robot velocity, fingertip positions, cube pose, cube velocity, and perturbation forces.
\begin{itemize}[leftmargin=1em]
    \item \textbf{Observations} (a) noisy estimates of the hand joint positions and velocities, (b) joint position errors (commanded vs achieved), (c) noisy estimates of the cube pose (distance of the cube to the palm center and cube orientation error), and (d) the previous commanded joint positions.  
    \item \textbf{Actions} 16 relative joint positions.
\end{itemize}

\paragraph{Training Setup}  
To promote sim-to-real transfer, we apply domain randomization on friction, cube mass, joint offsets, motor friction, reflected inertia, and PD gains, as well as link masses and sensor noise. We also add 2\,cm positional and 0.1\,rad rotational noise to the cube pose. We conduct two main training phases. During the first 200\,M steps, we train without random pose injection and torque limits. We then perform a 100\,M-step fine-tuning stage in which we introduce random pose ``injections'' with a 0.1 probability to mimic ``freak-out'' moments in real pose estimation (e.g., due to occlusions) and impose torque limits to match the real hardware.

\subsubsection{System Identification and Domain Randomization}
The original simulation environment for sim-to-real transfer provided by the LEAP Hand \cite{shaw2023leaphand} does not include robust system identification and instead relies heavily on manual parameter tuning. To improve both the performance and transparency of the system, we performed system identification on the DYNAMIXEL servo actuator used in the hand.

The armature inertia (i.e., rotor inertia reflected through the gearbox) for each joint is:
$
I_\text{a} = k_\text{g}^2 I_\text{r},
$
where \( k_\text{g} = 288.35 \) is the gear ratio from the supplier's data sheet, and
$
I_\text{r} = \frac{1}{2} m_\text{r} r_\text{r}^2 = \SI{1.7e-8}{\kilogram\meter\squared}
$
is the rotor inertia. To obtain \(I_\text{r}\), we assumed a uniform mass distribution of the rotor, based on physical disassembly and measurements of the rotor mass (\(m_\text{r} = \SI{2.0e-3}{\kilogram}\)) and radius (\(r_\text{r} = \SI{4.12e-3}{\meter}\)).

Because accurately modeling and measuring friction losses is difficult, we set 10\% of the maximum torque as the nominal friction value, and employed heavy domain randomization to account for uncertainties. 

For training, the servo actuator was controlled using a PD mapping similar to the locomotion setup in~\eqref{eq:pd_mapping}. However, during real-world deployment, the control law running on the servo actuator is: $i = k_\text{p}^\text{m} \bigl(\theta_\text{des}^\text{m} - \theta^\text{m}\bigr)
\;-\;
k_\text{d}^\text{m} \dot{\theta}^\text{m}$,
where \(i\) is the motor current command, and \(k_\text{p}^\text{m}, k_\text{d}^\text{m}, \theta_\text{des}^\text{m}, \theta^\text{m}\) are expressed in units different from those used in training. To reconcile these discrepancies, we assume \(\tau = k_\text{t} i\) and carefully compute the mappings based on the motor specifications provided in the data sheet.

Unlike the locomotion setup, the DYNAMIXEL actuator does not perform true current control (i.e., no direct motor current feedback). As a result, the above PD controller may deviate from the ideal behavior. To mitigate this mismatch, we introduce randomization in \(k_\text{p}\) and \(k_\text{d}\) parameters during training.

\subsubsection{Real Robot Setup}
We deploy the learned policy on the hand using its open-source software, with the following modifications:
\begin{itemize}[leftmargin=1em]
    \item \textbf{Control Frequency.} We reduce the policy control frequency from 150\,Hz to 20\,Hz in both simulation and real-world deployment, due to jitter issues with the low-level USB driver at higher frequencies.
    \item \textbf{System Identification.} We use the same torque (current) limit, stiffness, and damping parameters in training, guided by the system identification results described above.
\end{itemize}

\subsubsection{Vision-based Pose Estimator}

We use the vision-based cube pose estimator from \citep{handa2023dextreme} in order to solve for the pose of the cube, although any equivalent method of obtaining the SE(3) camera-to-object transformation would work. Given the local-space 3D coordinates of the cube and the 2D keypoints from the pose estimator, we solve for the camera-to-world transformation. Using camera-to-hand intrinsics calibration, we can then find the hand-to-cube transformation which is used as input to the policy. We run the cube pose estimator at 15\,Hz. During manipulation, we observe some small jitters or missed detections, but generally it is stable. Despite having access to three cameras on the physical hardware setup (\cite{li2024_drop}, we elect to use only one for simplicity.

\clearpage

\subsection{Real-world Non-prehensile Block Reorientation with a Franka-Robotiq Arm}
\label{appendix:RealWorldNonprehensileReorientationPolicyPerformance}

In this section, we provide technical details for our block reorientation task on a real Franka Emika Panda robot with a Robotiq gripper, including the simulation environment, training process, robot hardware interface, and camera-based object pose estimation. Our approach enables reliably learning and deploying a policy for non-prehensile manipulation of a yoga block, requiring only a brief training time in simulation while allowing zero-shot transfer to the real robot.  

\subsubsection{Simulation Environment}
\label{appendix:RealWorldNonprehensileReorientationPolicyPerformance:sim-env}

\begin{figure}[t]
    \centering
    \includegraphics[width=0.4\linewidth]{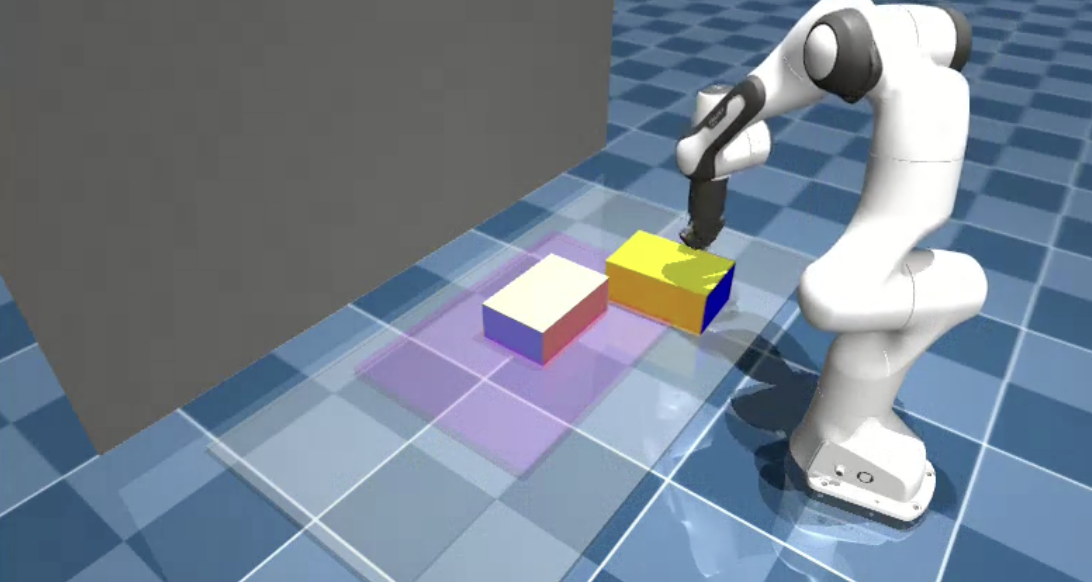}
    \caption{\small Example MuJoCo scene of our block reorientation environment. The block is pushed toward the center.}
    \label{fig:panda-robotiq-sim-env}
\end{figure}

The simulation environment (\Cref{fig:panda-robotiq-sim-env}) is designed to reorient a rectangular yoga block within a tabletop workspace region. The block is initialized at a random position and orientation subject to workspace bounds, and is then pushed, slid, or tapped to a desired goal pose at the center of the workspace. The policy uses 7D torque control signals for the robot arm and a fixed, closed Robotiq gripper. We include a simple termination condition when the block leaves the workspace or the end-effector violates safety constraints (e.g., collides with walls or floors).

\paragraph{Key Features}
\begin{itemize}[leftmargin=1em]
    \item \textbf{High-frequency torque control at 200\,Hz.} Each simulation step is advanced at a high frequency to match the targeted real-world controller rate.  
    \item \textbf{Curriculum learning.} We randomize initial joint positions, block poses, latencies in actions and observations, and other environment factors. A progressive curriculum increases the difficulty by gradually expanding the block's displacement and orientation range.  
    \item \textbf{Observation Delay.} Both actions and observations are delayed by random amounts at each episode step to approximate real hardware latencies.
    \item \textbf{Reward Shaping.} Shaped rewards encourage the robot to (i) stay near a nominal joint configuration, (ii) minimize velocities, (iii) keep the end-effector near the block, (iv) push the block toward the goal, and (v) orient the block to the desired angle.  
\end{itemize}

\paragraph{Policy Inputs and Actions} As shown in the environment code:
\begin{itemize}[leftmargin=1em]
    \item \textbf{Observations} (a) noisy estimates of the block pose, (b) current and recent robot joint positions and velocities, (c) the estimated end-effector pose, and (d) the target block pose.  
    \item \textbf{Actions} are 7D torque commands applied at the robot's joints. A constant action for the gripper (fingers closed) is appended for technical reasons in MuJoCo but remains fixed at a configured grasp.
\end{itemize}

\paragraph{Simulated Environment Details}  
\begin{itemize}[leftmargin=1em]
    \item \textbf{Gravity Compensation and Torque Bounds.} We configure the MuJoCo model to match the real robot's gravity compensation mode. Torque bounds are set to $8\,\mathrm{Nm}$ per joint in simulation, reflecting the approximate safe torque limit on real hardware.  
    \item \textbf{Collision Geometry.} The environment enforces collisions with floor, walls, and the block. The Robotiq gripper is held fixed but included for contact modeling.  
    \item \textbf{Delayed Observations and Actions.} We adopt random delays (between 1 and 3 steps for actions, and 6 to 12 steps for observations) to emulate real system communication latencies and sensor delay, following best practices in sim-to-real transfer.  
\end{itemize}

\paragraph{Training Setup}  
We train the policy on a 16x NVIDIA A100 GPU set-up for ten minutes of wall-clock time, with 3000 steps per episode (with action repeat set to 4, effectively running 750 policy decisions per episode). During training, the block's pose, robot states, and delays are heavily randomized. The final policy was selected from a checkpoint that achieved the highest success rate in the simulator.  

\subsubsection{Real Robot Setup}
\label{appendix:RealWorldNonprehensileReorientationPolicyPerformance:real-env}

\begin{figure}[t]
    \centering
    \includegraphics[width=0.8\linewidth]{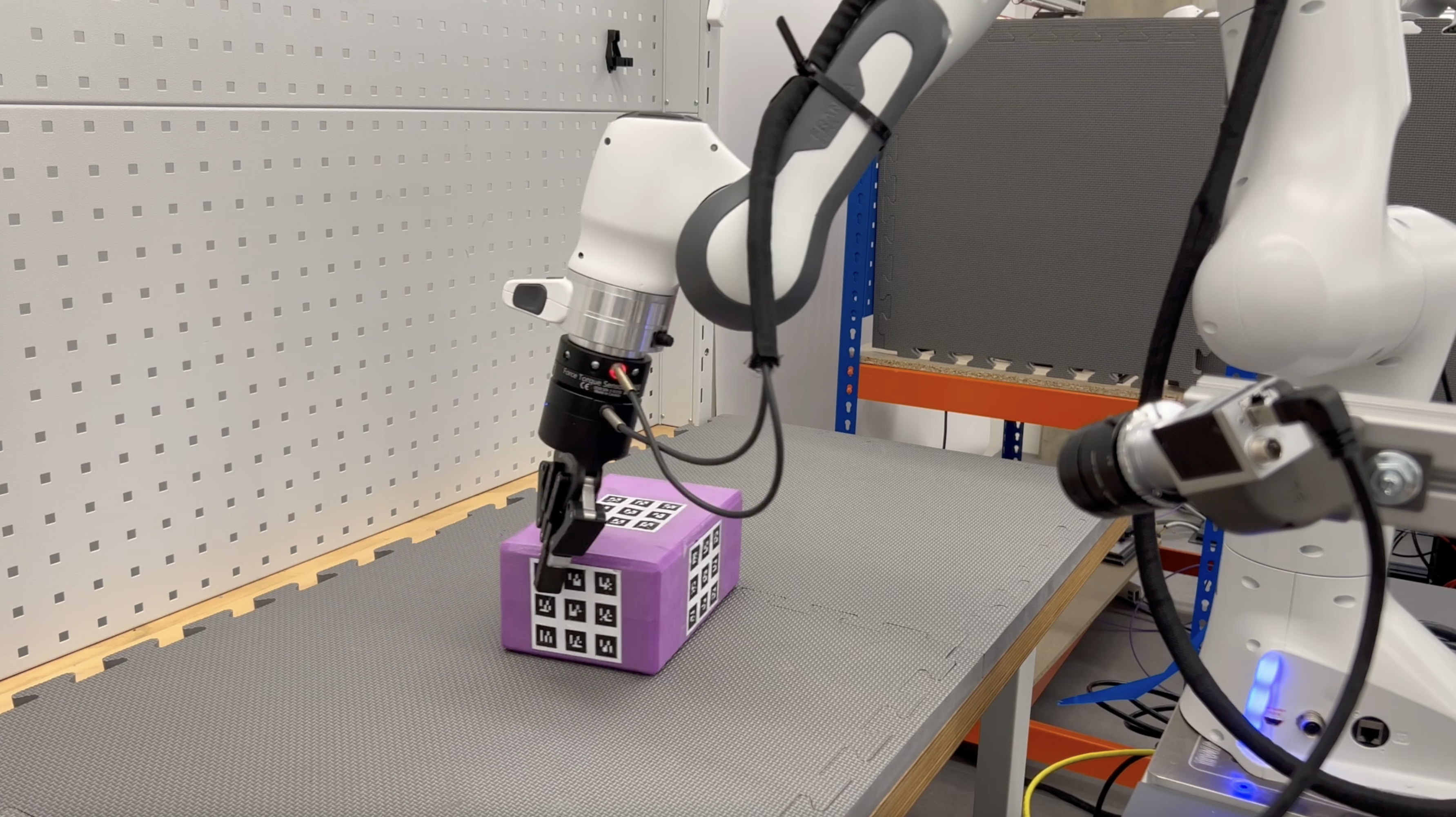}
    \caption{\small Real Franka Emika Panda robot with a Robotiq gripper, pushing the yoga block to the goal region.}
    \label{fig:panda-real-setup}
\end{figure}

We deployed the final trained policy on a Franka Emika Panda manipulator equipped with a Robotiq 2F-85 gripper and an integrated force torque sensor (Robotiq FT-300). Figure~\ref{fig:panda-real-setup} illustrates the hardware platform used for our experiments.

\paragraph{Direct Torque Control with Franka FCI}  
We interface with the robot via the Franka Control Interface (FCI) and send torque commands at 200\,Hz:
\begin{itemize}[leftmargin=1em]
    \item \textbf{Gravity Compensation.} The Panda is configured to compensate for the arm's own weight. The policy torques therefore focus on regulating the contact interactions with the block, making the system compliant.  
    \item \textbf{Bypassing Low-level PID Gains.} We avoid additional position or velocity tracking by sending raw joint torques. This significantly reduces the overhead of tuning any gain schedules and allows the learned policy to directly control contact forces.  
    \item \textbf{Safety Considerations.} We define software torque limits and monitor the robot's built-in safety stops and collision detection thresholds. In practice, the learned policy operates well within these limits to gently push the block.  
\end{itemize}

\paragraph{Control and Communication Pipeline}  
We use a lightweight C++/ROS node that relays torque commands to the Franka FCI at 200\,Hz:
\begin{itemize}[leftmargin=1em]
    \item \textbf{Policy Node in Python.} Our Python node loads the final trained policy (JIT-compiled for inference speed). At each 5\,ms tick, it receives the robot’s current joint positions, velocities, and the estimated block pose from ROS topics.  
    \item \textbf{Torque Message Publication.} The Python node computes a new 7D torque vector and publishes it as a ROS message to the C++ node. This node directly invokes the FCI’s real-time interface to set joint torques.  
    \item \textbf{Timing Synchronization.} We maintain a fixed 200\,Hz loop, matching the simulator’s update frequency. This avoids aliasing or missed steps and ensures that delays in the real system resemble the random delays already modeled in simulation.  
\end{itemize}

\subsubsection{Camera-based Block Pose Estimation}

The policy requires an estimate of the block’s 6D pose (position and orientation). We implement a multi-camera setup with four commodity RGB cameras:
\begin{itemize}[leftmargin=1em]
    \item \textbf{Intrinsics and Extrinsics.} Each camera is calibrated via OpenCV’s standard calibration procedure. We record images of a checkerboard pattern from various viewpoints to obtain precise intrinsic parameters (focal length, principal point) and extrinsic transformations.  
    \item \textbf{AR Tag Tracking.} We attach an Alvar~\citep{artrackalvar} fiducial marker to each face of the yoga block. Each camera runs the Alvar pose estimation pipeline. The final block pose is computed as the uniform average of valid detections.  
    \item \textbf{Placement Recommendations.} To improve coverage and reduce occlusions, we place two cameras at a lower height (approximately 40\,cm above the table) and two cameras overhead (around 80\,cm), all aimed toward the center of the workspace, inline with the base of the arm. This diversity of vantage points helps maintain robust tracking, even as the block is manipulated.  
    \item \textbf{ROS Integration.} Each camera node publishes pose estimates (with timestamps). A central ROS node fuses these estimates and broadcasts the block pose as a \texttt{geometry\_msgs/PoseStamped} message at about 30--60\,Hz.  
\end{itemize}

\paragraph{Summary} With this environment and training protocol, policies learned in simulation (under domain randomization and fast torque-control loops) exhibit a robust ability to transfer zero-shot to real hardware. Additionally, we encountered several limitations with the policy and workspace. For example, the policy sometimes pushed the block outside the robot’s workspace, making it impossible for the robot to reach it. We also observed that early versions of the policy moved the block too quickly, exceeding the robot’s force limit and causing it to pause. To address this, we introduced torque penalties, enabling the robot to maintain similar behavior while minimizing force. In summary we found that minimal engineering overhead was needed to align the MuJoCo-based environment with the real robot’s dynamic properties, underscoring the effectiveness of torque-based sim-to-real strategies with MuJoCo Playground.

\clearpage

\subsection{Real-world Franka PickCube from Pixels}
\label{appendix:RealWorldPixels}
To highlight the sim-to-real viability of our pixel-based environment, we highlight a robust real-world transfer using the Franka PickCube task.

\textbf{Task Description.} In our PickCube task, the goal of the robot is to move and grasp a 2x2x3 cm upright cube and return it to a fixed target position. To enable robust deployment with a single RGB camera, we limit both the object randomization and the robot's action space to a fixed Y-Z plane. We set the target in simulation to be (x, 0.0, 20.0), where x is set such that both the cube and the gripper initialize are in the same plane. Success is defined in simulation as lifting the object to a target height of 17 cm, and in real experiments as a stable grasp followed by lifting the object at least 10 cm above the table. The object's starting position is randomized along the Y-axis within a 20 cm range centered around 0. Because we train with randomized camera pose and a black background, we lay white tape over the range of possible cube starting positions to allow the memory-less policy to gauge its progress from the grasping site to the target height.

\begin{figure}[t]
    \centering
    \includegraphics[width=0.5\linewidth]{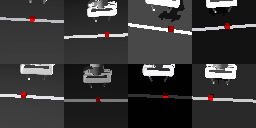}
    \caption{\small Policy inputs across domain randomized environments (64x64 pixels each) used while training the deployed PandaPickCubeCartesian agent. Lighting conditions, colors, brightness and camera pose are all randomized.}
    \label{fig:madrona-highres-domain-rando}
\end{figure}

\textbf{Training.} We use a similar reward shaping scheme as \cite{petrenko2023dexPBT}, using sparse rewards to encourage lifting the cube and bringing it close to the target position and dense rewards to guide the policy search in between. To simplify reward tuning, the dense reward terms only consider progress: $r_t = \mbox{clip}\left(\sum_i r_{t,i} - \max(r_1, r_2, ..., r_{t-1}), 0\right)$. This helps to emphasize the sparse terms during training. To improve sample efficiency, we terminate the policy upon completion. We train with randomized lighting conditions, colors, brightness and camera pose for robust real-world transfer as shown in \Cref{fig:madrona-highres-domain-rando}. Similar to the non-prehensile task in Section \ref{appendix:RealWorldNonprehensileReorientationPolicyPerformance:sim-env}, we adopt a random delay of 0 to 5 steps for the gripper action, as the real system has a small delay before the grippers begin to close. With an environment step of 50 ms, this results in the agent learning to adapt to a action delay of up to 0.25 s. We find the resulting conservative grasp behaviour to be important for sim-to-real transfer. We disable all except the pair-wise collisions between the gripper fingers and cube to increase simulation throughput.

\begin{figure}[t]
    \centering
    \includegraphics[width=0.8\linewidth]{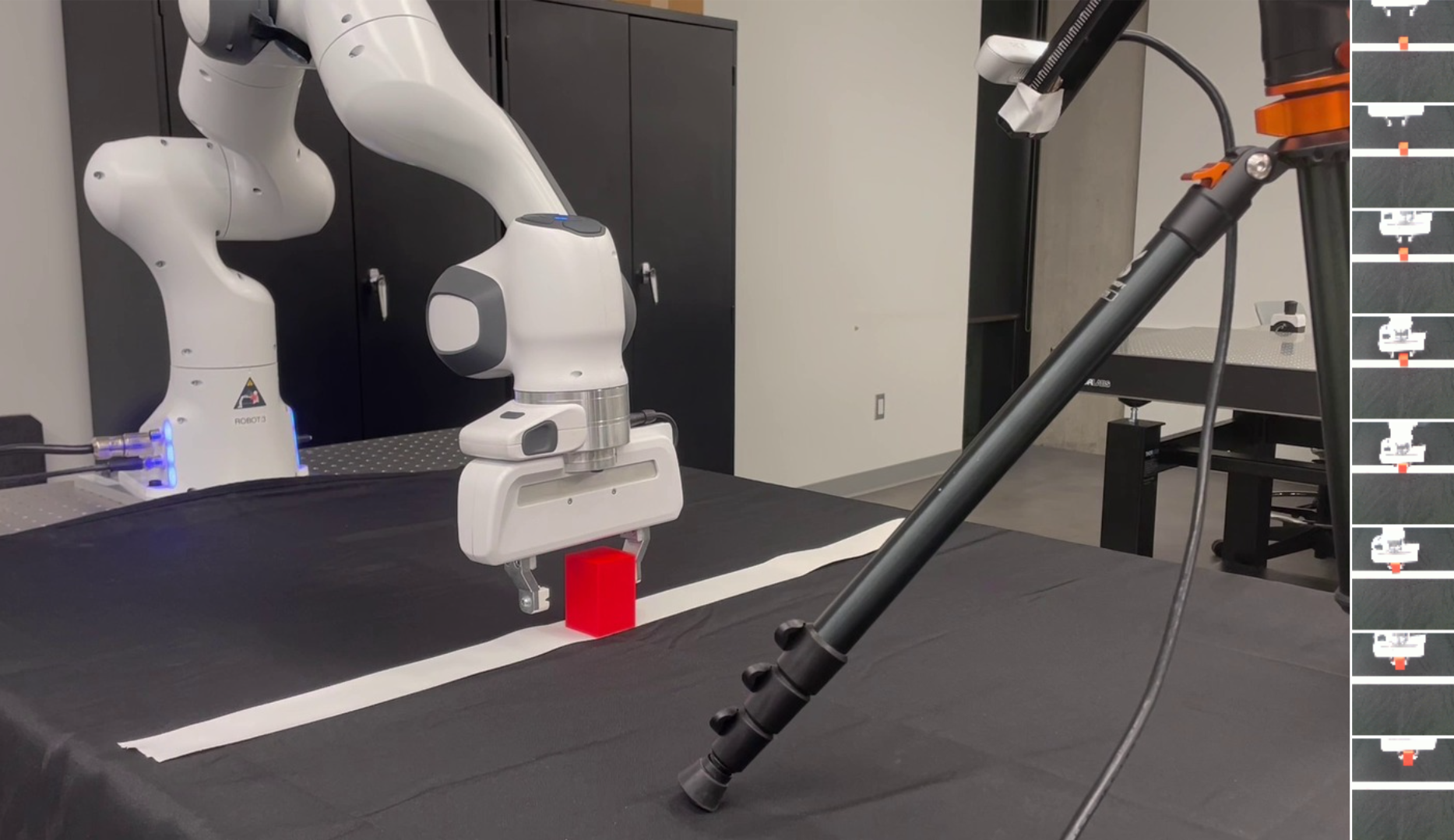}
    \caption{\small \textbf{Left.} Franka Research robot with a Realsense camera capturing input images. \textbf{Right.} Policy inputs from one embodied rollout.}
    \label{fig:panda-pixels-real-setup}
\end{figure}

\textbf{Agent.} Both the agent and critic networks comprise of a standard lightweight CNN architecture \cite{mnih2015human} followed by two hidden dense layers with size 256. Each channel of the input RGB image is individually normalised per sample by subtracting its mean and dividing it by its standard deviation. The policy network outputs a 3 value action from a single RGB camera looking down towards the gripper (Figure \ref{fig:panda-pixels-real-setup}). The first two actions are Cartesian increments in the Y and Z directions that is subsequently solved by an inverse kinematics controller \cite{he2021analytical}. X movement is ignored so that the gripper is restricted to the vertical plane of the block. We discretize the third action dimension to command a closed gripper when the policy value is below zero and an open position when greater than or equal to zero. All values are outputted in the range -1 to 1.

\textbf{Hardware.} We train our deployed policies within ten minutes on a single consumer-grade RTX 4090 GPU paired with a i9-14900KF processor. See \Cref{sec:madrona_training_curves} for training curves for the PandaPickCubeCartesian environment. We deploy on a Franka Research arm with an Intel D435 Realsense camera, using an RTX 3090 GPU for policy inference running at 15Hz.

\textbf{Control and Communication Pipeline.} We use a C++/ROS stack to execute our vision-based policies in real life. Camera images are square-cropped and down-sampled to 64x64 pixels before being passed to a lightweight C++ ONNX ROS Node for inference to produce a Cartesian increment and gripper command. This command is passed to a C++ ROS Node that computes joint commands using the same IK solution as used for training. These commands are output to a final ROS Node that wraps the Franka Control Interface (FCI) to control the robot joints. The control loop runs at 15 Hz, set by the incoming camera stream. We find that sim2real performance drastically improves from roughly calibrating the Cartesian increment scale in our physical setup to the one that the policy was trained on.

\clearpage

%% file: appendix/madrona.tex
\section{Madrona Rendering Environments}
\label{appendix:MadronaMJXBenchmark}

\subsection{RL Training Results}
\label{sec:madrona_training_curves}

\begin{figure}[ht]
    \centering
    \includegraphics[width=1.0\textwidth]{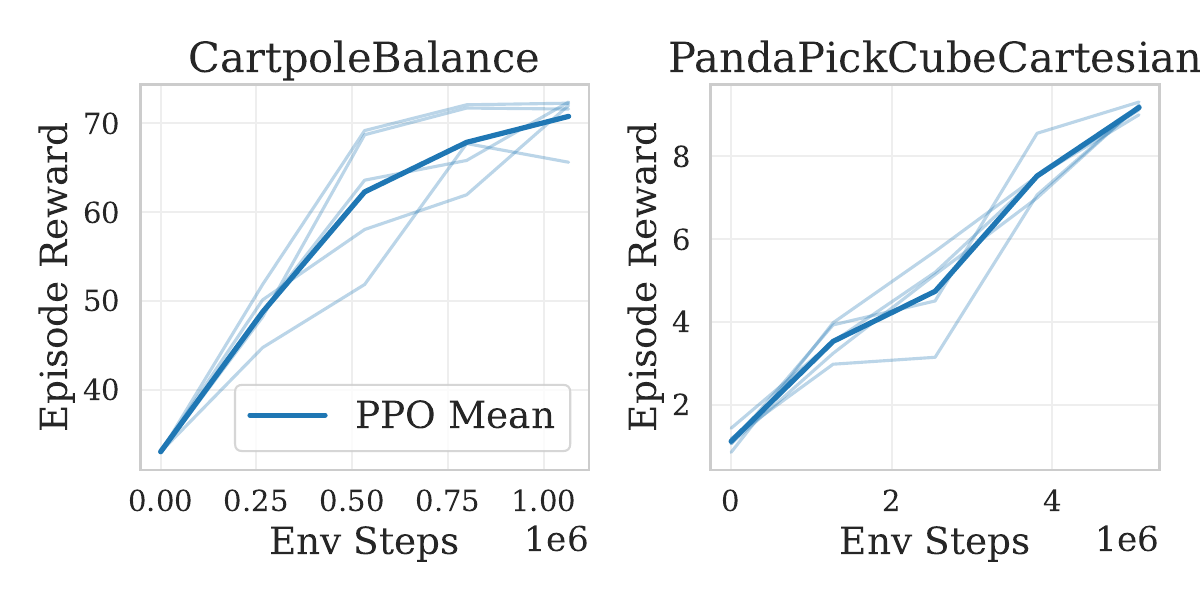}
    \caption{\small Reward vs environment steps for brax PPO. All settings are run with 5 seeds on a single RTX 4090 GPU.}
    \label{fig:madrona_step_reward}
\end{figure}

\begin{figure}[ht]
    \centering
    \includegraphics[width=1.0\textwidth]{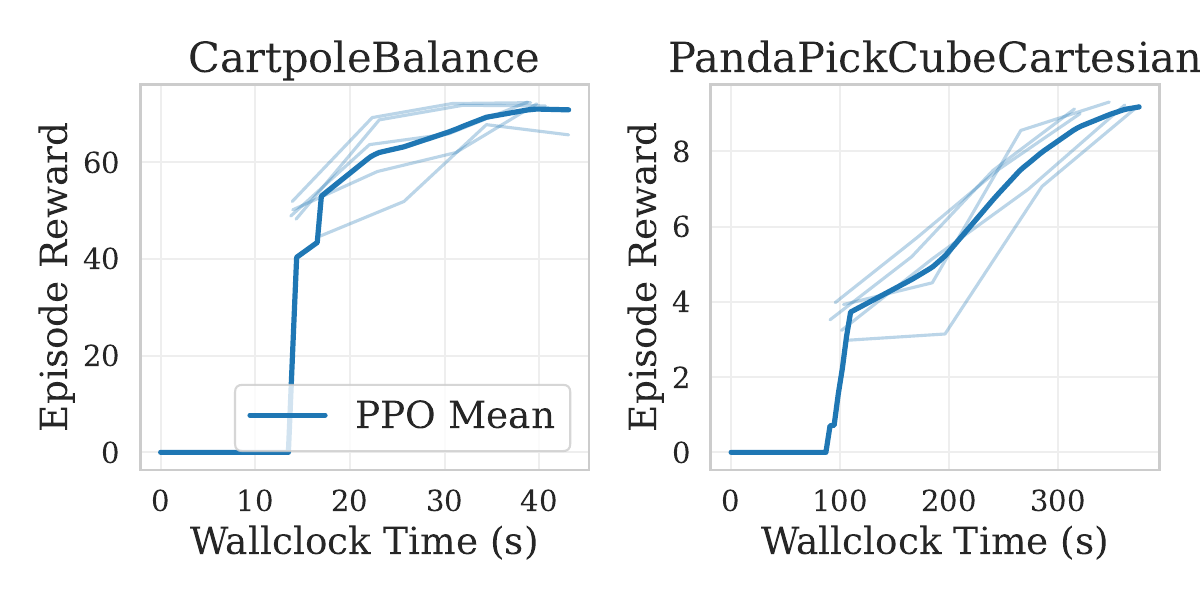}
    \caption{\small Reward vs wallclock time for brax PPO. All settings are run with 5 seeds on a single RTX 4090 GPU.}
    \label{fig:madrona_time_reward}
\end{figure}

MuJoCo Playground showcases two pixel observation environments using batch rendering; CartPoleBalance and PandaPickCubeCartesian. These two environments include complete training examples using brax PPO. We show the PPO training curves for both environments in Figures \ref{fig:madrona_step_reward} and \ref{fig:madrona_time_reward} across 5 seeds.

CartPoleBalance is adjusted for pixel-based observations by decreasing the control frequency such that more simulated experience can be factored into training with less policy updates, and by re-adjusting the RL training hyperparameters as necessary. The observations are of dimension 64x64x3 and consist of the current and previous two rendered observations, collapsed into grayscale then transposed for CNN inference. PandaPickCubeCartesian is derived from PandaPickCube. Our changes for faster and more stable pixel-based training are described in \Cref{appendix:RealWorldPixels}.

\subsection{Performance Benchmarking}
\label{sec:madrona_benchmarking}

\begin{figure}[t]
    \centering
    \includegraphics[width=0.85\linewidth]{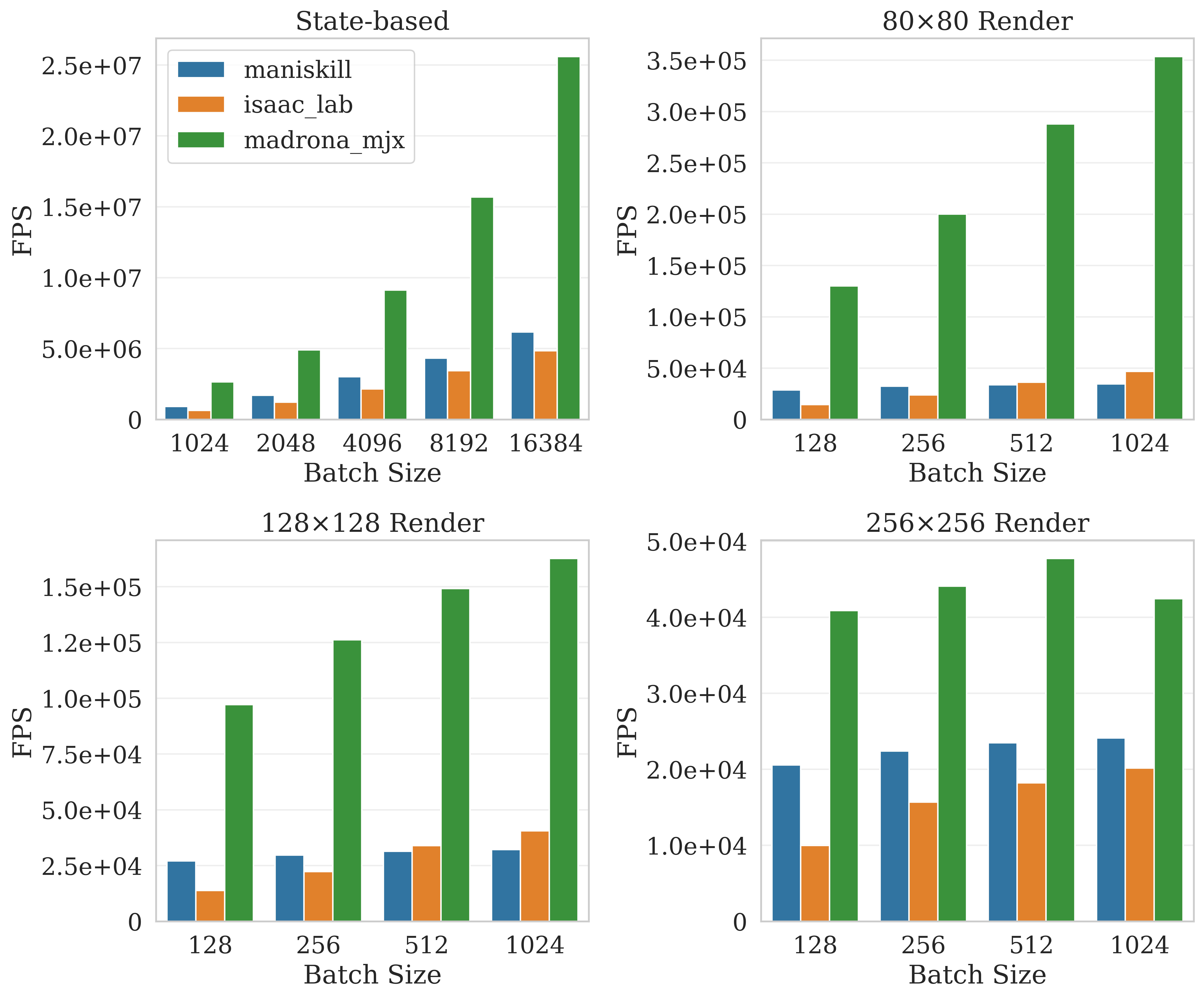}
    \caption{\small Comparison of raw environment-stepping throughput with prior simulators for CartpoleBalance with state-based and pixel observations of varying sizes.}
    \label{fig:cartpole_benchmark_combined}
\end{figure}

\begin{figure}[t]
    \centering
    \includegraphics[width=0.85\linewidth]{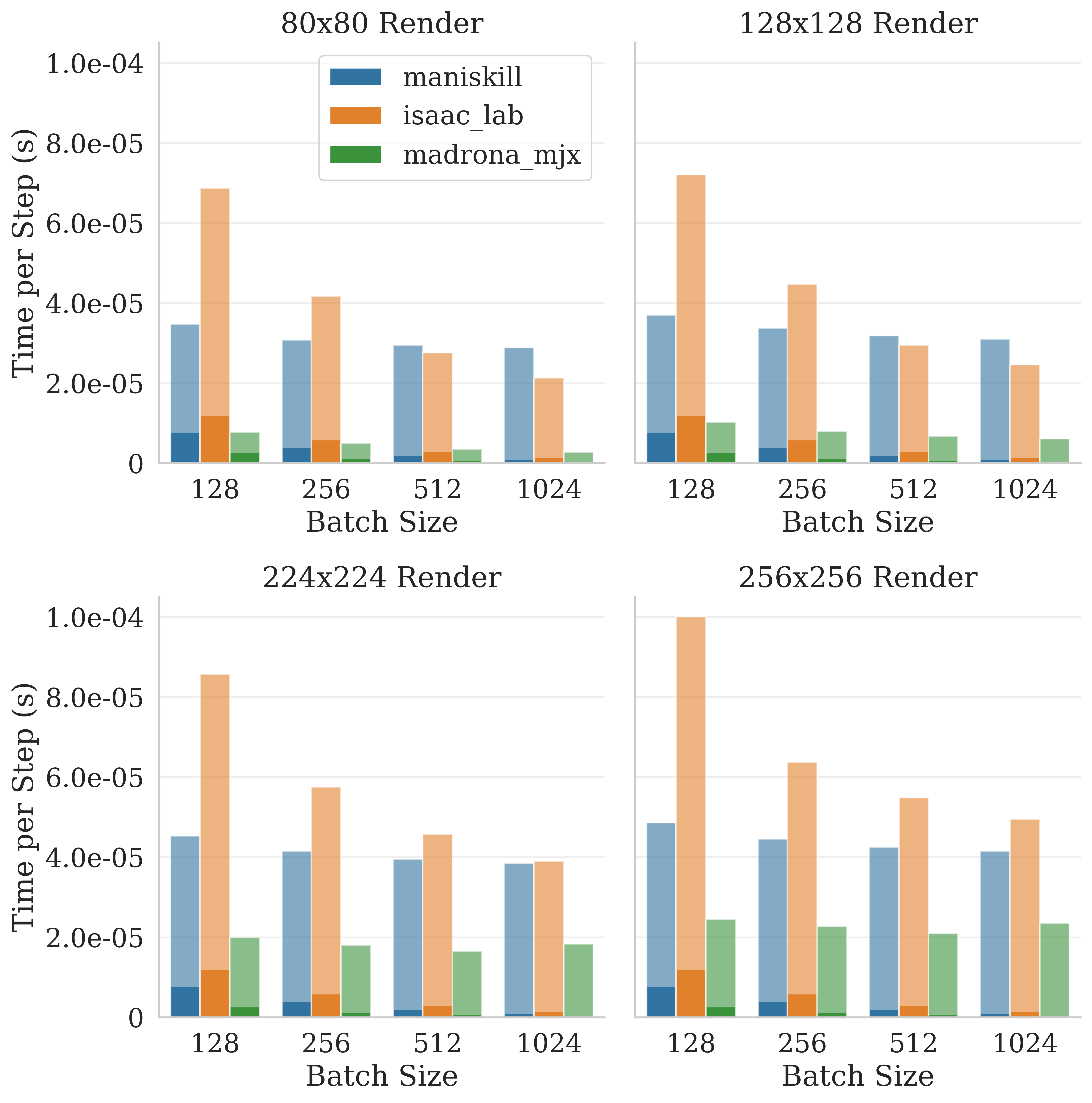}
    \caption{\small Time-cost breakdown of unrolling physics simulation and rendering for CartpoleBalance with pixel observations. \emph{Lower is better.} Per-step  rendering time is stacked without overlap over physics simulation time.}
    \label{fig:cartpole_benchmark_stacked}
\end{figure}

In this section we benchmark the througput of Madrona MJX GPU batch rendering. For reference, we plot our results alongside those from IsaacLab \cite{mittal2023orbit} and Maniskill3 \cite{tao2024maniskill3}; data for IsaacLab and Maniskill3 is obtained from \cite{tao2024maniskill3}. This is only a rough comparison as we only take steps to ensure similar hardware and timestep size, as a fully controlled performance benchmark is difficult due to the inherent differences between simulators. Our goal in these comparisons is to only highlight that our batch rendering is competitive with other state-of-the-art simulators that also include high-throughput rendering. %

The y-axis of Figure \ref{fig:cartpole_benchmark_combined} measures the rate of generating environment transitions $(s_t, o_t, r_t, s_{t+1})$ with random actions, comprising the basic data unit of most on and off-policy training algorithms. The first subplot measures Cartpole simulation with computationally trivial state-based observations. The next three plots show the cost of generating transitions where $o_t$ involves rendering with increasing resolution.

Figure \ref{fig:cartpole_benchmark_stacked} evaluates how much of our throughput increase is due to MJX's faster physics step. For each bar, the dark area shows the cost of the physics step and the non-overlapping light area shows the cost of generating the pixel observation. Note that lower values are better, as we display the inverse of frequency. While MJX's faster physics simulation indeed benefits throughput at lower image resolutions, Madrona's rendering speed improvements appear to be the primary driver of the measured speed-ups.

\clearpage

\subsection{Bottlenecks in Pixels-based Training}
\label{sec:appendix_madrona_bottlenecks}

\begin{table}[htbp]
    \centering
    \begin{tabular}{l c c c c}
    \toprule
             & \textbf{Env Step} & \textbf{with Pixels} & \textbf{and Inference} & \textbf{and Training} \\
    \midrule
    \multicolumn{5}{l}{\textbf{CartpoleBalance}} \\ 
    FPS         & \(1.37 \times 10^6\) & \(4.03 \times 10^5\) & \(3.41 \times 10^5\) & \(3.13 \times 10^4\) \\
    Time/Env Step (s) & \(7.30 \times 10^{-7}\) & \(2.48 \times 10^{-6}\) & \(2.93 \times 10^{-6}\) & \(3.20 \times 10^{-5}\) \\
    \midrule
    \multicolumn{5}{l}{\textbf{PandaPickCubeCartesian}} \\ 
    FPS         & \(6.40 \times 10^4\) & \(3.69 \times 10^4\) & \(3.60 \times 10^4\) & \(1.56 \times 10^4\) \\
    Time/Env Step (s) & \(1.56 \times 10^{-5}\) & \(2.71 \times 10^{-5}\) & \(2.78 \times 10^{-5}\) & \(6.39 \times 10^{-5}\) \\
    \bottomrule
    \end{tabular}
    \caption{Raw throughput of our two pixel-based environments in various settings. \emph{Env step}: stepping the physics with random actions. \emph{with Pixels}: Same, with the overhead of rendering pixel-based observations. \emph{with Inference}: random actions are replaced with policy inference. \emph{and Training}: the speed of PPO training. Results averaged over 5 runs on an RTX4090.}
    \label{tab:madrona_training_breakdown_measured}
\end{table}

\begin{table}[htbp]
    \centering
    \begin{tabular}{l c c c c}
    \toprule
             & \textbf{Physics} & \textbf{Rendering} & \textbf{Inference} & \textbf{Policy Update} \\
    \midrule
    \multicolumn{5}{l}{\textbf{CartpoleBalance}} \\ 
    Time/Env Step  (s)   & \(7.30 \times 10^{-7}\) & \(1.75 \times 10^{-6}\) & \(4.49 \times 10^{-7}\) & \(2.91 \times 10^{-5}\) \\
    Fraction & 0.02 & 0.06 & 0.01 & 0.91 \\
    \midrule
    \multicolumn{5}{l}{\textbf{PandaPickCubeCartesian}} \\ 
    Time/Env Step (s)        & \(1.56 \times 10^{-5}\) & \(1.15 \times 10^{-5}\) & \(6.45 \times 10^{-7}\) & \(3.62 \times 10^{-5}\) \\
    Fraction & 0.24 & 0.18 & 0.01 & 0.57 \\
    \bottomrule
    \end{tabular}
    \caption{Breakdown of total training time by component for CartpoleBalance and PandaPickCubeCartesian tasks, derived from Table \ref{tab:madrona_training_breakdown_measured}. Results averaged over 5 runs on an RTX4090.}
    \label{tab:madrona_training_breakdown_calc}
\end{table}

\Cref{tab:madrona_training_breakdown_calc} isolates the contributions of policy rollout (physics simulation, rendering, inference) and policy update to the overall cost per step in a training loop. We amortize the cost of policy update per policy rollout step, setting $t_4 = t_{training} + t_{inference} + t_{rendering} + t_{env step}$, where $t_4$ corresponds to \emph{Time/Env Step} in the last column of Table \ref{tab:madrona_training_breakdown_measured}. Working in reverse order through the table, we isolate each component. For example, $t_{training} = t_4 - t_3$ corresponds to the Policy Update \emph{Time/Env Step}.

We see that in both of our provided pixel-based environments, the training speed bottleneck is shifted from rendering to policy updates. This is especially true for the Cartpole, as the policy and value architectures includes convolutions determined by the size of the input image regardless of robot and task complexity. The expensive architecture coupled with the trivial embodiment, shift over 90\% of the training burden to network updates. For the Franka Panda environment, we buffer more of the computation into the physics by training with a lower control frequency. At 20 Hz control with a 5ms physics timestep, the policy makes only one decision per 10 simulator sub-steps. Similar to Cartpole, rendering is less of a bottleneck than the cost of processing the resultant images via convolutional-based network architectures.

\clearpage

%% file: appendix/rl_details.tex
\section{Reinforcement Learning Hyper-parameters}
\label{sec:rl_hypers}

In this section, we report the hyper-parameters used to train RL policies for all environments in MuJoCo Playground.

\subsection{DM Control Suite}

\begin{table}[ht]
\centering
\begin{tabular}{|l|c|p{5cm}|} 
\hline
\textbf{Hyperparameter} & \textbf{Default Value} & \textbf{Environment-Specific Modifications} \\ \hline
num\_timesteps & 60,000,000 & AcrobotSwingup, Swimmer, WalkerRun: 100,000,000 \\ \hline
num\_evals & 10 &  \\ \hline
reward\_scaling & 10.0 &  \\ \hline
normalize\_observations & True &  \\ \hline
action\_repeat & 1 & PendulumSwingUp: 4 \\ \hline
unroll\_length & 30 &  \\ \hline
num\_minibatches & 32 &  \\ \hline
num\_updates\_per\_batch & 16 & PendulumSwingUp: 4 \\ \hline
discounting & 0.995 & BallInCup: 0.95, \newline FingerSpin: 0.95 \\ \hline
learning\_rate & 1e-3 &  \\ \hline
entropy\_cost & 1e-2 &  \\ \hline
num\_envs & 2048 &  \\ \hline
batch\_size & 1024 &  \\ \hline
\end{tabular}
\caption{Brax PPO hyperparameters.}
\end{table}

\begin{table}[ht]
\centering
\begin{tabular}{|l|c|}
\hline
\textbf{Hyperparameter} & \textbf{Default Value} \\ \hline
madrona\_backend & True  \\ \hline
wrap\_env & False  \\ \hline
num\_timesteps & 1,000,000  \\ \hline
num\_evals & 5  \\ \hline
reward\_scaling & 0.1 \\ \hline
normalize\_observations & True  \\ \hline
action\_repeat & 1  \\ \hline
unroll\_length & 10  \\ \hline
num\_minibatches & 8  \\ \hline
num\_updates\_per\_batch & 8  \\ \hline
discounting & 0.97  \\ \hline
learning\_rate & 5e-4  \\ \hline
entropy\_cost & 5e-3  \\ \hline
num\_envs & 1024  \\ \hline
num\_eval\_envs & 1024  \\ \hline
batch\_size & 256  \\ \hline
\end{tabular}
\caption{Brax PPO hyperparameters for vision-based environments.}
\end{table}

\begin{table}[ht]
\centering
\begin{tabular}{|l|c|p{5cm}|} 
\hline
\textbf{Hyperparameter} & \textbf{Default Value} & \textbf{Environment-Specific Modifications} \\ \hline
num\_timesteps & 5,000,000 & Acrobot, Swimmer, Finger, Hopper, \newline CheetahRun, HumanoidWalk, \newline PendulumSwingUp, WalkerRun: 10,000,000 \\ \hline
num\_evals & 10 &  \\ \hline
reward\_scaling & 1.0 &  \\ \hline
normalize\_observations & True &  \\ \hline
action\_repeat & 1 & PendulumSwingUp: 4 \\ \hline
discounting & 0.99 &  \\ \hline
learning\_rate & 1e-3 &  \\ \hline
num\_envs & 128 &  \\ \hline
batch\_size & 512 &  \\ \hline
grad\_updates\_per\_step & 8 &  \\ \hline
max\_replay\_size & 1048576 * 4 &  \\ \hline
min\_replay\_size & 8192 &  \\ \hline
network\_factory.q\_network\_layer\_norm & True &  \\ \hline
\end{tabular}
\caption{Brax SAC hyperparameters.}
\end{table}

\clearpage

\subsection{Locomotion}

\begin{table}[ht]
\centering
\begin{tabular}{|l|c|} 
\hline
\textbf{Hyperparameter} & \textbf{Default Value} \\ \hline
num\_timesteps & 100,000,000 \\ \hline
num\_evals & 10  \\ \hline
reward\_scaling & 1.0 \\ \hline
normalize\_observations & True \\ \hline
action\_repeat & 1 \\ \hline
unroll\_length & 20 \\ \hline
num\_minibatches & 32 \\ \hline
num\_updates\_per\_batch & 4 \\ \hline
discounting & 0.97 \\ \hline
learning\_rate & 3e-4 \\ \hline
entropy\_cost & 1e-2 \\ \hline
num\_envs & 8192 \\ \hline
batch\_size & 256 \\ \hline
max\_grad\_norm & 1.0 \\ \hline
policy\_hidden\_layer\_sizes & (128, 128, 128, 128) \\ \hline 
policy\_obs\_key & "state" \\ \hline
value\_obs\_key & "state" \\ \hline
\end{tabular}
\caption{Default Brax PPO hyperparameters.}
\end{table}

\begin{table}[ht]
\centering
\begin{tabular}{|l|c|} 
\hline
\textbf{Hyperparameter} & \textbf{Value} \\ \hline
num\_timesteps & 200,000,000 \\ \hline
num\_evals & 10  \\ \hline
num\_resets\_per\_eval & 1 \\ \hline
policy\_hidden\_layer\_sizes & (512, 256, 128) \\ \hline 
value\_hidden\_layer\_sizes & (512, 256, 128) \\ \hline 
value\_obs\_key & "privileged\_state" \\ \hline
\end{tabular}
\caption{Brax PPO hyperparameters specific to Go1JoystickFlatTerrain and Go1JoystickRoughTerrain.}
\end{table}

\begin{table}[ht]
\centering
\begin{tabular}{|l|c|} 
\hline
\textbf{Hyperparameter} & \textbf{Value} \\ \hline
num\_timesteps & 100,000,000 \\ \hline
num\_evals & 5  \\ \hline
policy\_hidden\_layer\_sizes & (512, 256, 128) \\ \hline 
value\_hidden\_layer\_sizes & (512, 256, 128) \\ \hline 
value\_obs\_key & "privileged\_state" \\ \hline
\end{tabular}
\caption{Brax PPO hyperparameters specific to Go1Handstand and Go1Footstand.}
\end{table}

\begin{table}[ht]
\centering
\begin{tabular}{|l|c|} 
\hline
\textbf{Hyperparameter} & \textbf{Value} \\ \hline
num\_timesteps & 200,000,000 \\ \hline
num\_evals & 10  \\ \hline
discounting & 0.95 \\ \hline
policy\_hidden\_layer\_sizes & (512, 256, 128) \\ \hline 
value\_hidden\_layer\_sizes & (512, 256, 128) \\ \hline 
value\_obs\_key & "privileged\_state" \\ \hline
\end{tabular}
\caption{Brax PPO hyperparameters specific to Go1Backflip.}
\end{table}

\begin{table}[ht]
\centering
\begin{tabular}{|l|c|} 
\hline
\textbf{Hyperparameter} & \textbf{Value} \\ \hline
num\_timesteps & 50,000,000 \\ \hline
num\_evals & 5  \\ \hline
policy\_hidden\_layer\_sizes & (512, 256, 128) \\ \hline 
value\_hidden\_layer\_sizes & (512, 256, 128) \\ \hline 
value\_obs\_key & "privileged\_state" \\ \hline
\end{tabular}
\caption{Brax PPO hyperparameters specific to Go1Getup.}
\end{table}

\begin{table}[ht]
\centering
\begin{tabular}{|l|c|} 
\hline
\textbf{Hyperparameter} & \textbf{Value} \\ \hline
num\_timesteps & 400,000,000 \\ \hline
num\_evals & 16  \\ \hline
num\_resets\_per\_eval & 1 \\ \hline
reward\_scaling & 0.1 \\ \hline
unroll\_length & 32 \\ \hline
num\_updates\_per\_batch & 5 \\ \hline
discounting & 0.98 \\ \hline
learning\_rate & 1e-4 \\ \hline
entropy\_cost & 0 \\ \hline
num\_envs & 32768 \\ \hline
batch\_size & 1024 \\ \hline
clipping\_epsilon & 0.2 \\ \hline
policy\_hidden\_layer\_sizes & (512, 256, 64) \\ \hline 
value\_hidden\_layer\_sizes & (256, 256, 256, 256) \\ \hline 
value\_obs\_key & "privileged\_state" \\ \hline
\end{tabular}
\caption{Brax PPO hyperparameters specific to G1Joystick.}
\end{table}

\begin{table}[ht]
\centering
\begin{tabular}{|l|c|} 
\hline
\textbf{Hyperparameter} & \textbf{Value} \\ \hline
num\_timesteps & 400,000,000 \\ \hline
num\_evals & 16  \\ \hline
num\_resets\_per\_eval & 1 \\ \hline
reward\_scaling & 0.1 \\ \hline
unroll\_length & 32 \\ \hline
num\_updates\_per\_batch & 5 \\ \hline
discounting & 0.98 \\ \hline
learning\_rate & 1e-4 \\ \hline
entropy\_cost & 0 \\ \hline
num\_envs & 32768 \\ \hline
batch\_size & 1024 \\ \hline
clipping\_epsilon & 0.2 \\ \hline
policy\_hidden\_layer\_sizes & (512, 256, 64) \\ \hline 
value\_hidden\_layer\_sizes & (256, 256, 256, 256) \\ \hline 
value\_obs\_key & "privileged\_state" \\ \hline
\end{tabular}
\caption{Brax PPO hyperparameters specific to T1Joystick.}
\end{table}

\begin{table}[ht]
\centering
\begin{tabular}{|l|c|} 
\hline
\textbf{Hyperparameter} & \textbf{Value} \\ \hline
num\_timesteps & 100,000,000 \\ \hline
num\_evals & 10  \\ \hline
num\_resets\_per\_eval & 1 \\ \hline
clipping\_epsilon & 0.2 \\ \hline
discounting & 0.99 \\ \hline
learning\_rate & 1e-4 \\ \hline
entropy\_cost & 0.005 \\ \hline
policy\_hidden\_layer\_sizes & (512, 256, 128) \\ \hline 
value\_hidden\_layer\_sizes & (512, 256, 128) \\ \hline 
value\_obs\_key & "privileged\_state" \\ \hline
\end{tabular}
\caption{Brax PPO hyperparameters specific to Berkeley Humanoid.}
\end{table}

\clearpage

\subsection{Manipulation}

\begin{table}[ht]
\centering
\begin{tabular}{|l|c|} 
\hline
\textbf{Hyperparameter} & \textbf{Default Value} \\ \hline
normalize\_observations & True \\ \hline
reward\_scaling & 1.0 \\ \hline
policy\_hidden\_layer\_sizes & (32, 32, 32, 32) \\ \hline 
policy\_obs\_key & "state" \\ \hline
value\_obs\_key & "state" \\ \hline
\end{tabular}
\caption{Default Brax PPO hyperparameters.}
\end{table}

\begin{table}[ht]
\centering
\begin{tabular}{|l|c|} 
\hline
\textbf{Hyperparameter} & \textbf{Value} \\ \hline
num\_timesteps & 150,000,000 \\ \hline
num\_evals & 10 \\ \hline
unroll\_length & 40 \\ \hline
num\_minibatches & 32 \\ \hline
num\_updates\_per\_batch & 8 \\ \hline
discounting & 0.97 \\ \hline
learning\_rate & 3e-4 \\ \hline
entropy\_cost & 1e-2 \\ \hline
num\_envs & 1024 \\ \hline
batch\_size & 512 \\ \hline
policy\_hidden\_layer\_sizes & (256, 256, 256, 256) \\ \hline 
\end{tabular}
\caption{Brax PPO hyperparameters for AlohaSinglePegInsertion.}
\end{table}

\begin{table}[ht]
\centering
\begin{tabular}{|l|c|} 
\hline
\textbf{Hyperparameter} & \textbf{Value} \\ \hline
num\_timesteps & 40,000,000 \\ \hline
num\_evals & 4 \\ \hline
unroll\_length & 10 \\ \hline
num\_minibatches & 32 \\ \hline
num\_updates\_per\_batch & 8 \\ \hline
discounting & 0.97 \\ \hline
learning\_rate & 1e-3 \\ \hline
entropy\_cost & 2e-2 \\ \hline
num\_envs & 2048 \\ \hline
batch\_size & 512 \\ \hline
policy\_hidden\_layer\_sizes & (32, 32, 32, 32) \\ \hline 
num\_resets\_per\_eval & 1 \\ \hline
\end{tabular}
\caption{Brax PPO hyperparameters for PandaOpenCabinet.}
\end{table}

\begin{table}[ht]
\centering
\begin{tabular}{|l|c|} 
\hline
\textbf{Hyperparameter} & \textbf{Value} \\ \hline
num\_timesteps & 5,000,000 \\ \hline
num\_evals & 5 \\ \hline
unroll\_length & 10 \\ \hline
num\_minibatches & 8 \\ \hline
num\_updates\_per\_batch & 8 \\ \hline
discounting & 0.97 \\ \hline
learning\_rate & 5.0e-4 \\ \hline
entropy\_cost & 7.5e-3 \\ \hline
num\_envs & 1024 \\ \hline
batch\_size & 256 \\ \hline
reward\_scaling & 0.1 \\ \hline
policy\_hidden\_layer\_sizes & (256, 256) \\ \hline 
num\_resets\_per\_eval & 1 \\ \hline
max\_grad\_norm & 1.0 \\ \hline
\end{tabular}
\caption{Brax PPO hyperparameters for PandaPickCubeCartesian.}
\end{table}

\begin{table}[ht]
\centering
\begin{tabular}{|l|c|} 
\hline
\textbf{Hyperparameter} & \textbf{Value} \\ \hline
num\_timesteps & 20,000,000 \\ \hline
num\_evals & 4 \\ \hline
unroll\_length & 10 \\ \hline
num\_minibatches & 32 \\ \hline
num\_updates\_per\_batch & 8 \\ \hline
discounting & 0.97 \\ \hline
learning\_rate & 1e-3 \\ \hline
entropy\_cost & 2e-2 \\ \hline
num\_envs & 2048 \\ \hline
batch\_size & 512 \\ \hline
policy\_hidden\_layer\_sizes & (32, 32, 32, 32) \\ \hline 
\end{tabular}
\caption{Brax PPO hyperparameters for PandaPickCube.}
\end{table}

\begin{table}[ht]
\centering
\begin{tabular}{|l|c|} 
\hline
\textbf{Hyperparameter} & \textbf{Value} \\ \hline
num\_timesteps & 2,000,000,000 \\ \hline
num\_evals & 10 \\ \hline
unroll\_length & 100 \\ \hline
num\_minibatches & 32 \\ \hline
num\_updates\_per\_batch & 8 \\ \hline
discounting & 0.994 \\ \hline
learning\_rate & 6e-4 \\ \hline
entropy\_cost & 1e-2 \\ \hline
num\_envs & 8192 \\ \hline
batch\_size & 512 \\ \hline
num\_resets\_per\_eval & 1 \\ \hline
num\_eval\_envs & 32 \\ \hline
policy\_hidden\_layer\_sizes & (64, 64, 64, 64) \\ \hline 
\end{tabular}
\caption{Brax PPO hyperparameters for PandaRobotiqPushCube.}
\end{table}

\begin{table}[ht]
\centering
\begin{tabular}{|l|c|} 
\hline
\textbf{Hyperparameter} & \textbf{Value} \\ \hline
num\_timesteps & 100,000,000 \\ \hline
num\_evals & 10 \\ \hline
num\_minibatches & 32 \\ \hline
unroll\_length & 40 \\ \hline
num\_updates\_per\_batch & 4 \\ \hline
discounting & 0.97 \\ \hline
learning\_rate & 3e-4 \\ \hline
entropy\_cost & 1e-2 \\ \hline
num\_envs & 8192 \\ \hline
batch\_size & 256 \\ \hline
num\_resets\_per\_eval & 1 \\ \hline
policy\_hidden\_layer\_sizes & (512, 256, 128) \\ \hline 
value\_hidden\_layer\_sizes & (512, 256, 128) \\ \hline 
policy\_obs\_key & "state" \\ \hline
value\_obs\_key & "privileged\_state" \\ \hline
\end{tabular}
\caption{Brax PPO hyperparameters for LeapCubeRotateZAxis).}
\end{table}

\begin{table}[ht]
\centering
\begin{tabular}{|l|c|} 
\hline
\textbf{Hyperparameter} & \textbf{Value} \\ \hline
num\_timesteps & 100,000,000 \\ \hline
num\_evals & 20 \\ \hline
num\_minibatches & 32 \\ \hline
unroll\_length & 40 \\ \hline
num\_updates\_per\_batch & 4 \\ \hline
discounting & 0.99 \\ \hline
learning\_rate & 3e-4 \\ \hline
entropy\_cost & 1e-2 \\ \hline
num\_envs & 8192 \\ \hline
batch\_size & 256 \\ \hline
num\_resets\_per\_eval & 1 \\ \hline
policy\_hidden\_layer\_sizes & (512, 256, 128) \\ \hline 
value\_hidden\_layer\_sizes & (512, 256, 128) \\ \hline 
policy\_obs\_key & "state" \\ \hline
value\_obs\_key & "privileged\_state" \\ \hline
\end{tabular}
\caption{Brax PPO hyperparameters for LeapCubeReorient.}
\end{table}

\begin{table}[ht]
\centering
\begin{tabular}{|l|c|} 
\hline
\textbf{Hyperparameter} & \textbf{Value} \\ \hline
madrona\_backend & True \\ \hline
wrap\_env & False \\ \hline
normalize\_observations & True \\ \hline
reward\_scaling & 1.0 \\ \hline
policy\_hidden\_layer\_sizes & (32, 32, 32, 32) \\ \hline 
num\_timesteps & 5,000,000 \\ \hline
num\_evals & 5 \\ \hline
unroll\_length & 10 \\ \hline
num\_minibatches & 8 \\ \hline
num\_updates\_per\_batch & 8 \\ \hline
discounting & 0.97 \\ \hline
learning\_rate & 5.0e-4 \\ \hline
entropy\_cost & 7.5e-3 \\ \hline
num\_envs & 1024 \\ \hline
batch\_size & 256 \\ \hline
reward\_scaling & 0.1 \\ \hline
num\_resets\_per\_eval & 1 \\ \hline
\end{tabular}
\caption{Brax PPO hyperparameters for vision-based PandaPickCubeCartesian.}
\end{table}